\documentclass[conference]{IEEEtran}

\IEEEoverridecommandlockouts
\usepackage{cite}
\usepackage{amsmath,amssymb,amsfonts}
\usepackage{algorithm}
\usepackage{algpseudocode}
\usepackage{graphicx}
\usepackage{float}
\usepackage{textcomp}
\usepackage{xcolor}
\usepackage{multirow}
\usepackage{booktabs}
\usepackage{tabularx}
\usepackage{subcaption}
\usepackage{multicol}

\newcommand\floor[1]{\lfloor#1\rfloor}
\newcommand\ceil[1]{\lceil#1\rceil}

\def\BibTeX{{\rm B\kern-.05em{\sc i\kern-.025em b}\kern-.08em
    T\kern-.1667em\lower.7ex\hbox{E}\kern-.125emX}}
\begin{document}

\title{Feature Importance Guided Attack, a Model Agnostic
Adversarial Attack for Tabular Datasets}

\author{\IEEEauthorblockN{Gilad Gressel} 
\IEEEauthorblockA{\textit{Center for Cybersecurity} \\
Amrita University, India \\
gilad.gressel@am.amrita.edu}\\

\IEEEauthorblockN{Kalyani Harikumar }
\IEEEauthorblockA{\textit{Center for Cybersecurity} \\
Amrita University, India}\\
\\

\and
\IEEEauthorblockN{Niranjan Hegde}
\IEEEauthorblockA{\textit{Center for Cybersecurity} \\
Amrita University, India}\\
\\

\IEEEauthorblockN{Anjali S.}
\IEEEauthorblockA{\textit{Center for Cybersecurity} \\
Amrita University, India}

\and
\IEEEauthorblockN{Archana Sreekumar}
\IEEEauthorblockA{\textit{Center for Cybersecurity} \\
Amrita University, India}\\
\\

\IEEEauthorblockN{Krishnashree Achuthan}
\IEEEauthorblockA{\textit{Center for Cybersecurity} \\
Amrita University, India}

\and
\IEEEauthorblockN{Rishikumar Radhakrishnan}
\IEEEauthorblockA{\textit{Center for Cybersecurity}\\
Amrita University, India}

}
\maketitle
\thispagestyle{plain}
\pagestyle{plain}

\begin{abstract}

Research in adversarial learning has primarily focused on homogeneous unstructured datasets, which often map into the problem space naturally. Inverting a feature space attack on heterogeneous datasets into the problem space is much more challenging, particularly the task of finding the perturbation to perform. This work presents a formal search strategy: the `Feature Importance Guided Attack' (FIGA), which finds perturbations in the feature space of heterogeneous tabular datasets to produce evasion attacks. We first demonstrate FIGA in the feature space and then in the problem space.
FIGA assumes no prior knowledge of the defending model's learning algorithm and does not require any gradient information. FIGA assumes knowledge of the feature representation and the mean feature values of defending model's dataset. FIGA leverages feature importance rankings by perturbing the most important features of the input in the direction of the target class. While FIGA is conceptually similar to other work which uses feature selection processes (e.g., mimicry attacks), we formalize an attack algorithm with three tunable parameters and investigate the strength of FIGA on tabular datasets. We demonstrate the effectiveness of FIGA by evading phishing detection models trained on four different tabular phishing datasets and one financial dataset with an average success rate of 94\%. We extend FIGA to the phishing problem space by limiting the possible perturbations to be valid and feasible in the phishing domain. We generate valid adversarial phishing sites that are visually identical to their unperturbed counterpart and use them to attack six tabular ML models achieving a 13.05\% average success rate. 

\end{abstract}

\begin{IEEEkeywords}
Phishing Detection, Adversarial Attacks, Machine Learning, Problem Space Attack
\end{IEEEkeywords}

\section{Introduction}

Adversarial attacks on machine learning pose a significant threat to the security of all machine learning applications. In recent years a significant amount of research has gone into studying and designing adversarial attacks and defenses on machine learning systems. Most of this research has been on homogeneous and unstructured datasets, such as computer vision and natural language processing. However, many commonly performed ML tasks, such as phishing detection, malware detection, recommender systems, credit review, fraud detection, and healthcare assessments, rely on heterogeneous tabular datasets. 

In a homogeneous dataset, all features are semantically identical, and the feature space often directly maps to the problem space. For example, in an image dataset, all the features (e.g., pixels) are continuous and bounded with the same range (e.g., 0-255). Most of the literature has focused on deep neural networks, encouraging the use of gradient-based attacks applied directly to the feature space. Attacks on homogeneous datasets performed in feature space (e.g., on the pixels) directly produce feasible and valid examples in problem space (e.g., recognizable images). However, the attacks that work on homogeneous datasets cannot be applied straightforwardly to heterogeneous datasets~\cite{mathov_not_2021}. 

Heterogeneous or tabular datasets pose challenges because their features are not standardized, contain missing values, and have categorical, nominal, or continuous feature types. Tabular features may have certain restrictions to maintain validity, e.g., a person's age cannot be negative. Some features may be immutable and cannot be changed (e.g., a phishing site cannot obtain www.google.com as its domain name). When searching for an adversarial perturbation, the most popular gradient-based attacks do not readily apply in the context of non-ordered categorical and nominal features. Finally, unlike homogeneous datasets, perturbation of the features may not produce a valid adversarial example in problem space. In tabular datasets, there is rarely a direct mapping from feature space to problem space~\cite{pierazzi_intriguing_2020}. For example, suppose we perturb a phishing website with additional images and links to mimic a benign website. In that case, these additions will also increase other features, such as the amount of text and the number of href tags. These unexpected changes are known as side effect features and cannot easily be accounted for.

While there have been successful attacks against tabular datasets in feature and problem space, the attack strategies employed are ad-hoc compared to the plethora of homogeneous attack algorithms which can be applied to any homogeneous dataset. Adversarial attacks against tabular datasets have been deployed since the early 2000s, beginning with attacks on IDS and malware ~\cite{fogla_evading_nodate,fogla_polymorphic_nodate, biggio_evasion_2013}. These were mimicry attacks; they perturbed the input to mimic the target. Recent research on tabular attacks proposes novel attacks that attempt generic attacks on tabular datasets. However they contain flaws such as dropping all categorical features~\cite{ballet_imperceptible_2019}, are expensive to run due to genetic algorithms ~\cite{hashemi_permuteattack_2020}, or employ complex surrogate functions ~\cite{mathov_not_2021}. While Pierazzi et al. introduce a formalization for adversarial ML evasion attacks in the problem space and demonstrate methods to evaluate different problem space attacks, they do not propose any new search strategies to find the perturbations~\cite{pierazzi_intriguing_2020}.

In this work, we build on the intuitions behind mimicry attacks and develop an evasion attack against heterogeneous tabular datasets: the `Feature Importance Guided Attack' (FIGA). FIGA perturbs the $n$ most important features (chosen by a ranking algorithm) of the input in the direction of the target class by $\epsilon$ percent. FIGA has three parameters: $n$, the total number of features to perturb, $\epsilon$, the amount of perturbation to apply, and $f_i$, a feature importance ranking algorithm. FIGA perturbs the $\epsilon$ amount of feature value to all $n$ features \emph{in the direction} of the target class. For example, consider URLs representing phishing and benign websites. The average length of a benign URL is longer than the average length of a phishing URL. FIGA would perturb the phishing URLs to become longer in order to appear more legitimate. FIGA is model agnostic: it does not require knowledge of the defending model's algorithm or hyperparameters. FIGA requires the feature representation and the mean feature values of the dataset in order to perform its attack. For these reasons, FIGA is a limited-knowledge attack. We demonstrate FIGA both in feature space and in problem space.

In feature space, we use FIGA to generate adversarial perturbations and demonstrate their effectiveness by attacking five machine learning models: Random Forest, Multilayer Perceptron, Logistic Regression, XGBoost, and a FastAI tabular model. We trained all models on four different phishing datasets and one finance dataset. FIGA achieves an average 94\% success rate on all datasets and models. In problem space, we generate 10,000 adversarial phishing websites (visibly identical to their unperturbed counterparts), which evade the same models with an average success rate of 52\%. The performance loss from feature space to problem space is primarily due to the problem space constraints (certain features cannot be perturbed). We will present a detailed analysis of FIGA using security evaluation curves to examine the range of FIGA's attack strength across all three parameters $n, \epsilon$, and $fi$.

\begin{itemize}
     \item The creation of the `Feature Importance Guided Attack' (FIGA) a novel model-agnostic, gradient-free attack algorithm for tabular datasets.
     \item Demonstration of FIGA on five phishing detection models trained on four phishing datasets and the adult dataset.
     \item Creation of 10,000 adversarial phishing websites in problem space which can evade detection.
\end{itemize}

\section{Background and Related Work}

An adversarial evasion attack is any technique designed to mislead a machine learning model at inference time. Adversaries craft malicious `adversarial examples’ and feed them as input to a machine learning model..

\subsection{Adversarial Threat Models}
Adversarial attacks are categorized into three threat scenarios based on the attacker's capability: perfect knowledge white-box attacks, limited knowledge gray box attacks, and zero knowledge black box attacks. In the perfect knowledge model, the attacker has complete knowledge of the training data, algorithm, hyper-parameters, and learned parameters. In the limited knowledge model, the attacker has partial knowledge of the feature representation and the learning algorithm but not the learning algorithm's hyper-parameters or training data.~\cite{nazemi2019potential,sotgiu2019deep,hitaj2017deep}. In the zero-knowledge model, the attacker has no knowledge but is allowed to query an `oracle' representing the defending model. Additionally, it is assumed that the attack will have \emph{some} knowledge about the feature representation of the defending model~\cite{biggio_wild_2018}. Without any knowledge (either approximate or exact) of the feature representation or transformations to make on the input, it would be impossible to attack a model with zero knowledge.

FIGA is a limited knowledge attack since FIGA assumes knowledge of the feature representation and requires, at the minimum, the mean feature values of the dataset used by the defending model to mimic the target class. However, FIGA is completely model agnostic and gradient-free. It does not require knowledge of the target model hyper-parameters, architecture, or learned parameters.

\subsection{Attacks on Tabular Datasets}
Recently there have been several novel attacks on tabular datasets. Each attack has focused on solving the specific challenges of their domain. A few attacks focus primarily on financial datasets \cite{sarkar_robust_2018, hashemi_permuteattack_2020, cartella_adversarial_nodate}. Sarkar et al. propose a novel attack titled Max Salience Attack (MSA), which focuses on minimizing the number of features changed (only 1 or 2 features are allowed to be perturbed)\cite{sarkar_robust_2018}. This attack is similar to FIGA in that it causes a perturbation by a fixed percent change. However, their work differs because they only explore continuous features of the dataset. Hashemi et al. describe a genetic algorithm in which they randomly sample actual feature values from the dataset, thus ensuring that the resulting permutation is valid. Finally, Cartella et al. modify the loss function of ZOO and perform tabular attacks on the German dataset \cite{cartella_adversarial_nodate}. 

Ballet et al. create a general form adversarial attack for any tabular dataset~\cite{ballet_imperceptible_2019}. However, they only permute the continuous features of the dataset, dropping nominal features and treating all ordered categorical features as continuous. Their main contribution is the weighting of the features' importance (they argue that only the less important features should be perturbed - an argument we disagree with) and custom distance metrics to ascertain the imperceptibility of the attack.  

Finally, Mathov et al. introduce a novel attack that relies on a surrogate model to learn a gradient-based perturbation. Their essential idea is to create an embedding function $f$, which transforms the tabular features into a latent continuous space. They build a surrogate model in the latent space, attack it, and reverse the samples into the original space while maintaining data validity. The main drawback of their attack is the complexity and effort required.

\subsubsection{Data Validity}

Data validity encompasses two notions: data consistency and data feasibility. Consistency is the idea that specific perturbations of the target feature may result in samples that simply would not or could not occur in the domain. For example, a person's birth date cannot occur after their death date. The other key concept is data feasibility: in many real-world applications, certain data \emph{cannot be modified}. For example, it is infeasible for a phishing site to have the domain www.google.com (as this domain cannot be purchased). These two notions of consistency and feasibility are formalized by Mathov et al. in \cite{mathov_not_2021}, and we also apply these constraints to our work.

\subsubsection{Imperceptibility}
The notion of imperceptibility is that the adversarial sample should fool the model, not a human (it should look like the original input). In the context of images and audio is intuitive and can be readily ascertained with human judgment. Either the picture looks like the original, or it does not, but "looks" is defined by human perception, which collectively we tend to agree upon. However, imperceptibility is not intuitive or easily ascertained in the context of tabular heterogeneous datasets. In most tabular datasets, a human domain expert would be required to determine if the sample was imperceptible or not. Cartella et al. specify a custom norm that penalizes solutions that perturbed important features that would be checked by a human operator \cite{cartella_adversarial_nodate}. Erdemir et al. tackle this problem with the Mahalanobis distance and a weighted norm (using feature importance)~\cite{erdemir_adversarial_2021}. 

In our work, we attack phishing webpages that rely on visual similarity to the victim pages. Therefore as long as the perturbation could be applied to the source code of the input webpage with an invisible output on the webpage, we believe the attack will be imperceptible. We rely on the fact that all perturbations will translate to code modifications that can be made hidden in the source code but not displayed in the browser to the user. Further, when performing our attack in problem space, we only add features and do not subtract - this simplifying approach ensures that the original phishing website remains identical in functionality.

\subsection{Adversarial Phishing Attacks} Phishing is a common social engineering attack that exposes users' confidential data. Machine learning models can be trained to accurately detect phishing sites in real-time~\cite{vinayakumar2018evaluating,kiruthigaphishing,abu2007comparison,basnet2008detection}. These models use features extracted from a website's URL, HTML, and network attributes to detect whether or not the site is compromised or malicious~\cite{mohammad2014intelligent,yerima2020high,darling2015lexical}. Often these features are categorical, binary, and discrete.

Recent work has demonstrated successful adversarial attacks against phishing detection models. We observe three attack strategies: deep learning to modify URLs~\cite{10.1145/2976749.2978330, bahnsen2018deepphish}, other types of novel URL modification techniques~\cite{lei2020advanced, sabir2020evasion} and modification of visual features such as HSL (hue, saturation, light)~\cite{panum2020towards}.

Three studies perform evasion attacks by modifying the URL alone. Sabir et al. propose a technique similar to a homoglyph attack, which generates novel phishing URLs that evade detection~\cite{sabir2020evasion}. Bahnsen et al. created an algorithm `DeepPhish' which uses an LSTM neural network in order to create synthetic URLs~\cite{bahnsen2018deepphish}. DeepPhish aims to increase the effectiveness of phishing URLs by altering those already created by the attacker. Aleroud and Karabatis proposed using generative adversarial networks (GANs) to generate adversarial URL samples~\cite{10.1145/3375708.3380315}. They found that the adversarial samples generated with GANs were effective at evading models which use Intra-URL relationships.
All three of these approaches do not make use of HTML features. Thus, if a detection model uses HTML-based features, they may not evade detection. In contrast to these algorithms, FIGA can be applied to any arbitrary dataset provided a suitable feature ranking method exists. 

Lee et al. proposed a defense against an attack that replaces the values of the most important features of the phishing site with the corresponding values from a benign site~\cite{leeBuildingRobustPhishing2020}. The attack designed by Lee et al. showed that the AUPRC (area under the precision-recall curve) score of their phishing detection random forest model dropped from 99.1 to 69.4. We performed FIGA against the dataset by Lee et al., and the AUPRC score dropped from 96.1 to 39.3. The work done by Lee et al. focuses on the defense yet does not formalize or clearly explain their method of attack. They write that they copy features from the target class but do not mention details regarding the quantity and selection of the perturbations. Finally, the attack takes place in feature space, and it is quite likely that the attack would not be feasible in problem space.

Shirazi et al. propose an adversarial random sampling attack on phishing detection~\cite{shirazi_adversarial_2019}. They use publicly available datasets where every feature is binned to either -1, 0, or 1; where -1 is a legitimate instance, 0 is suspicious, and 1 indicates phishing. The attack algorithm consists of random feature replacement (using cartesian products) from the known phishing samples. In their later work, they used a clustering approach to direct their sampling attack \cite{shirazi_directed_nodate}. The main drawback is the limited datasets (40k samples in total) and no discussion about the feasibility of the newly constructed samples. In contrast, FIGA is a systematic attack guaranteed to step in the direction of the target class. Their attack is not mapped into the problem space.



\section{Methodology}

We will present the FIGA algorithm and the experimental results in feature and problem space. First, we will review the threat model and assumptions, followed by the datasets we used to validate our results. Then we will present the algorithm, followed by details of the experiments performed. Finally, we will analyze the results with security evaluation curves in both feature and problem space.

\subsection{Threat Model and Assumptions}

FIGA is a limited knowledge attack that assumes knowledge of the dataset and feature representation, but \emph{not} the learning algorithm or its hyper-parameters. In the case of phishing, we assume that the attacker has control over the URL and HTML code of the phishing page.

\begin{table}
\centering
\caption{Phishing class distribution for the primary dataset.}
\begin{tabular}{@{}lll@{}}
\toprule
                    & \textbf{Phishing} & \textbf{Legitimate} \\ \midrule
\textbf{samples} & 138,473            & 210,266              \\
\end{tabular}
\label{dataset_count}
\end{table}

\subsection{Data Collection}

We collected both phishing and benign website data with a Selenium-based web crawler. Selenium, an instrumented crawler, replicates user behavior. This is important because many phishing sites, to avoid detection, serve web crawlers false information~\cite{marchal2018designing,zeber2020representativeness}. We collected our phishing data using URLs taken from PhishTank~\cite{phishtank} throughout 2019-2020. Phishtank.com is widely used for ground-truth phishing labels since its submissions are manually verified by human users~\cite{marchal2016know,tian2018needle}. We collected our legitimate data using URLs taken from Tranco~\cite{LePochat2019} created on 19 February 2020\footnote{https://tranco-list.eu/list/433X/200000}. Tranco aggregates from four URL ranking lists: Alexa, Majestic, Umbrella, and Quantcast while removing any URLs flagged by Google safe browsing. Ranking sites are commonly used to collect benign websites in phishing research since the most trafficked websites will be the most scrutinized and, therefore, most likely benign~\cite{yang2020nature,varshney2016survey}. We crawled the benign website recursively, following links on each page to a maximum depth of five. This results in finding long, detailed URLs collected deeper in a website's structure. We collected 348,739 URLs and their associated HTML source code between January 2019 - March 2020. Table~\ref{dataset_count} shows the distribution of samples. This dataset henceforth will be referred to as the primary dataset. All experiments performed used 80\% of the data for training and 20\% for validation.

\subsection{Feature Extraction for the primary dataset}
We extracted 52 features from the website's URLs and HTML. Some features are the protocol used in a URL, the number of digits in a URL, the length of the subdomain, the count of meta tags, script tags, the length of the text in a page's body, and the number of images on a page. We choose these features since they are used commonly in the phishing literature~\cite{mohammad2012assessment,mohammad2015phishing}. All HTML and URL features are nearly entirely under the attacker's control. The only element out of an attacker's control is that the domain name must be unique and not already purchased by another party (e.g., an attacker cannot use `www.google.com' as their domain). Otherwise, the attacker is free to manipulate all elements of the website. The full feature set for the primary dataset can be found in the appendix. 

It should be noted that many features affect each other. For example, if we were to increase the number of javascript tags in the HTML, we would also implicitly increase the text in the body. These intertwined features are known as side-effects and increase the difficulty of converting successful perturbations in feature space to problem space~\cite{pierazzi_intriguing_2020}. When creating the actual HTML elements in problem space, we cause these side effects to other features.

\subsection{Alternate datasets} \label{alt}
Along with the primary dataset, we use four other datasets to test the effectiveness and robustness of FIGA. A 10,000 sample phishing dataset from \cite{kaggle_datset}(\emph{Kang}) a 15,000 sample dataset from \cite{iscx_dataset}(\emph{Mamun}) and the 130,000 sample phishing dataset proposed by Lee et al\cite{leeBuildingRobustPhishing2020}(\emph{Lee}). Finally, we attempted a single non-phishing dataset. We attacked the Adult dataset from the UCI Machine Learning Repository\cite{adult_dataset}(\emph{Adult})

\subsection{Feature Importance Guided Attack}
FIGA involves a two-step process: ranking the features according to their importance and the subsequent perturbation of those features found to be most significant.

\subsubsection{Ranking Features \& Discovering the Attack Direction}
The first step in FIGA is to rank the features and find the perturbation direction for each one. This process is seen in Algorithm~\ref{featureAlgo}.

\begin{algorithm}
	\caption{\label{featureAlgo}Feature Importance \& Attack Direction}
	 \hspace*{\algorithmicindent} \textbf{Input:} \text{$X,y,f_{i}$} \\
     \hspace*{\algorithmicindent} \textbf{Output:} \text{$f$, vector ranked features} \\
     \hspace*{\algorithmicindent} \hspace*{\algorithmicindent} \hspace*{\algorithmicindent} \text{$d$, attack direction vector}
     
	\begin{algorithmic}[1]
	    \State $f \gets$ $f_{i}$($X,y$)
        \State $d \gets sign \left[\textrm{mean}(f_{input})-\textrm{mean}(f_{target})\right]$
        
        \State \textbf{return} $f, d$ 
    \end{algorithmic} 
\end{algorithm}

\begin{table}
\caption{Feature importance methods used}
\centering
\begin{tabular}{@{}c@{}}
\textbf{Feature Importance Method}   \\ \midrule
Information Gain Ratio      \\
Gini Impurity Coefficient   \\
Permutation Importance      \\
Forward Feature Selection   \\
Recursive Feature Selection \\ \bottomrule
\end{tabular}
\label{feat_imp}
\end{table}

\begin{table}
\centering
\caption{10 Most Important Features (Gini Impurity) in the Primary Dataset}
\begin{tabular}{@{}clc@{}}
\toprule
\textbf{Rank} & \textbf{Feature}  & \textbf{Sign}   \\ 
\midrule
1 & \# href           & +               \\
2 & \# www    & +               \\
3 & \# dir in url   & -               \\
4 & \# digits in html         & -                \\
5 & \# javascript         & +               \\
6 &  len subdomain       & -               \\
7 & \# text in body          & +                \\
8 &  protocol          & +                \\
9 & \# meta          & +                \\
10 &  len free url          & -                \\

\bottomrule
\end{tabular}

\label{table:top_ten_feat}

\end{table}
As input, we require a dataset $X$ with labels $y$ and feature importance method $f_{i}$. Any feature importance method or ranking algorithm can be used in step 1. We tested FIGA with five feature importance methods in Table ~\ref{feat_imp}. \emph{Gini impurity coefficient} (Gini) and \emph{Information Gain} (IG) both measure the ability of a feature to split the dataset into pure subgroups. These algorithms are commonly used in decision tree algorithms to build the tree. \emph{Permutation} importance measures the change in performance after shuffling a feature column individually. If the difference is significant, then that feature is critical. \emph{Recursive Feature Elimination} (RFE) is a meta-ranking algorithm. It uses an estimator that can rank features (either through a univariate approach or coefficients of the learning algorithm) and then recursively drops the least performing feature. \emph{Forward feature selection} (FFS) starts with a null model and then starts fitting the model with each feature one at a time. In each iteration, a feature is added that best improves the model until adding a new variable does not improve the model's performance. While we tested these five methods, many more options could be used as a valid $fi$ method.

Step 2 returns a vector $d$ which contains the $sign$ (+/-) for each feature in the vector. For each feature, its $sign$ value determines the direction in which it will be perturbed. We calculate the $signs$ by examining the mean feature values of the target class (the class we would like to mimic) and then compare them with the input class. In our case, the input class is phishing. We increase or decrease a phishing example's feature values to make it appear more legitimate (the target class). For example, if the average length of the phishing URLs is longer than the mean size of legitimate URLs, we have a negative $sign$ and need to shorten the malicious ones. Table~\ref{table:top_ten_feat} shows the direction associated with the ten highest ranking features from the primary dataset using info gain as the ranking algorithm.



\subsubsection{Perturbation of Features}

Algorithm~\ref{perturb_algo} contains the steps required to perform the feature perturbations on an input sample.

\begin{algorithm}
	\caption{Perturbation Algorithm} \label{perturb_algo}
	 \hspace*{\algorithmicindent} \textbf{Input:} \text{$f, d, n, \epsilon, X_{train}, x, T$}\\
     \hspace*{\algorithmicindent} \textbf{Output:} \text{$x^*$}
	\begin{algorithmic}[1]
	    \State (min, max) $\gets T.fit(X_{train}$)
	    \State $x \gets T.transform(x$, min, max)
	    
	    \State $\epsilon \gets \frac{\epsilon}{n} * sum(x_{features}$) 
	    \For{$i \gets 0, n$} \Comment{iterate features to perturb}
        \State $x[i] \gets x[i] + (\epsilon * d[n])$
        \If {$x[i] <0$}
		 \State $x[i] \gets 0$
		\EndIf
		\If{$x[i] >1$} 
		 \State $x[i] \gets 1$
		\EndIf
        \EndFor
	    
		\State $x^* \gets T^{-1}(x)$
		\For{$i \gets 0, n$} \Comment{iterate perturbed features}
		\If{$i$ is discrete} \Comment{select discrete features}
		    \If {$d[i]$ is positive}
		    \State $x^*[i] \gets x^*\floor{i}$
		    \Else
		    \State $x^*[i] \gets x^*\ceil{i}$
		        \EndIf
		        \EndIf
		\EndFor
		\State \textbf{return} $x^*$
	\end{algorithmic} 
\end{algorithm} 

As inputs to Algorithm~\ref{perturb_algo} we have $f$ (a ranked feature vector), $d$ (a signed attack direction vector), $n$ (the number of features to perturb), $\epsilon$ (the total percentage of the input to modify), $X\_train$ (training data), $x$ (the sample we would like to perturb), and $T$ (a scaling transformer). The output is $x^*$, the adversarially perturbed feature set of the input sample. The two parameters that the user will be interested in selecting will be $\epsilon$ and $n$, both of which control the strength of the attack. 

Step 1 of the algorithm transforms the data with $T$, the Min\-Max transformer, which results in all feature values between 0 and 1. This is done so we can clip any resulting perturbation within the bounds of 0 and 1. This will maintain data consistency, and our new perturbed features are bounded by the min and max of the original features. In Step 3, we sum the phishing sample's feature values and multiply them by $\frac{\epsilon}{n}$, the desired perturbation percentage divided by the number of features to perturb. We divide the perturbation amount by $n$ to equally apply the perturbation across features. This yields an $\epsilon$ value that is now relative to the number of features we will perturb and the total feature magnitude of the input sample. 

In step 4, we begin the for loop to perturb each of the $n$ features in the sample $x$. Each feature is perturbed in proportion to its magnitude. We update each feature individually according to the feature directions found in Algorithm~\ref{featureAlgo}. Since we have min-maxed the dataset, we ensure that perturbed values are neither greater than one nor lesser than zero, thus keeping the values within the empirical constraints of the dataset.
In step 4 we begin the for loop to perturb each of the $n$ features in the sample $x$. Each feature is perturbed in proportion to its own magnitude. We update each feature individually according to the feature directions found in Algorithm~\ref{featureAlgo}. Since we have min-maxed the dataset, we ensure that perturbed values are neither greater then one nor lesser then zero thus keeping the values withing the empirical constraints of the dataset.

In step 13, we inverse transform the sample into the original feature space. We note that floating-point feature values may be introduced into the dataset on discrete columns. To address this, in step 14, we loop through the features, filter for the discrete ones, round down any feature which increased in size, and round up any feature which decreased in size. If the feature is discrete, we ensure a valid perturbation to ensure data consistency. We always round towards the original feature value, which guarantees that the perturbation is $<=\epsilon$: the desired sample (phishing site) perturbation percentage. 

Finally, it should be noted that in our experiments, we one-hot encoded all categorical features and treated them as discrete binary variables. In this way, we can use FIGA to find the direction of each categorical variable separately.

\subsubsection{Restrictions on Adversarial Examples in the Problem Space}

While it is simple to perturb any feature, it is not easy to always translate those perturbations back into problem space. This is known as the \emph{inverse feature-mapping problem}~\cite{pierazzi_intriguing_2020}. In order to create valid adversarial samples in the problem space, we restricted our feature selections to only HTML features with a positive direction that we could add to the source without affecting the page's visual appearance. We did not use any features with a negative direction as that would require removing parts of the website. We believe that a motivated attacker \emph{could and would} remove parts of their website to evade detection. However, to guarantee non-breaking changes through the automated generation of adversarial samples, we did not.

In our experiments, we found that we could craft perturbations that could be translated into live phishing pages. These features are listed in Table~\ref{table:prob_space_feat}. We added invisible HTML elements to perturb the page without affecting the appearance of the pages.

\section{Experiments}

\subsection{Classifiers to Attack}

We selected a total of six models to attack. We implemented Random Forests, Multi-layer Perceptron, Decision Tree and Logistic regression using Scikit-Learn~\cite{sklearn_api}. We used the Fastai~\cite{howard2018fastai} library to implement a deep learning model designed for tabular datasets. Finally, we tested with gradient boosted trees using XGBoost~\cite{Chen:2016:XST:2939672.2939785}.

\subsection{FIGA Hyperparameters} \label{hyper}
FIGA has three hyperparameters, the feature importance method $f_i$, the number of features $n$, and the amount of perturbation $\epsilon$, that control the strength and the effectiveness of the attack. To explore the strength of the FIGA attack, we searched exhaustively through the specified subset of all the parameter space for every dataset with each classifier. For example, for the primary dataset we varied $n$ from 1 to 52 ( the maximum number of features), $\epsilon$ was varied from 0.001 to 4.0 with 50 even steps while cycling between all the available feature importance methods. 

The decision to search $\epsilon$ to 4.0 is motivated by the fact that, on average, a legitimate website in our PD is 2.7 times the size of a phishing website. The average phishing website is 47.6 kilobytes, while the average size of a phishing website is 126.5 kilobytes. To mimic legitimate websites, we must grow the size of a phishing website. We searched beyond 2.7 to examine the FIGA attack's unbounded strength.

While a grid search is effective at finding the arguments which maximize the attack, it pays no attention to the cost of the attack. To this end, we plot success rate curves in section~\ref{succ_curves} to determine the best trade-off between attack strength and cost.

When converting to problem space, we searched a reduced subset of these parameters as the problem space attack is computationally expensive to perform due to creating the HTML source code and re-extracting the features from the resulting website. We used the feature space results to narrow the search of the problem space attack.
 

\section{Feature Space Results and Discussion}
In this section, we will discuss the results of our experiments in feature space. In feature space, we allowed ourselves to perturb all features (both adding or subtracting) without regard for plausibility in problem space. We perform these experiments for two reasons: it is computationally cheaper to search the feature space, and we wanted to understand the theoretical ability of the FIGA algorithm to evade classifiers. If FIGA could not produce adversarial samples in feature space, it would certainly not work in problem space. Our findings showed that FIGA can evade all models provided an unbounded $\epsilon$ and that we can also search the feature space to find valid problem space perturbations.
\begin{table*}
\centering
\caption{The results of grid-search FIGA for each model in feature space when attacking the primary dataset}
\begin{tabular}{lccccclc}
\toprule
Trained Models      & \multicolumn{2}{c}{Recall}  & Success Rate\% & $n$  & $\epsilon$     & $f_{i}$ \\ \toprule
                       & Baseline      & Attack  &    &    &       &                           \\ \hline
FastAI Tabular Model               & 91.7         & 0.000  & 100    & 49 & 0.491 & \emph{info gain}         \\ 
Multi Layer Perceptron             & 96.6         & 0.0  & 100    & 7  & 0.57 & \emph{info gain}                      \\
Logistic Regression                & 90.6         & 0.0  & 100     & 6  & 0.246 & \emph{info gain}                      \\
Decision Tree                      & 96.1         & 1.3  & 98.7    & 25 & 4.0 & \emph{permutation}                      \\
XGBoost                           & 97.0         &  0.4  & 99.5    & 20  & 4.0 & \emph{rfe}               \\
Random Forest                      & 97.5         & 3.7  & 96.2     & 21 & 4.0 & \emph{rfe}        \\ \bottomrule  

\end{tabular}

\label{Performance}
\end{table*}

\subsection{Metrics}
When measuring an adversarial attack's performance, it is appropriate to use the success rate. The success rate is the percentage of adversarially perturbed samples that evaded the model. The recall represents the percentage of phishing samples detected, and a perfect attack will have zero samples detected. In our experiments, we did not remove samples already misclassified by the model (false negatives) because we wanted to measure the adversarial attack on all samples in the off-hand chance that the perturbations improved the classifiers' performance. Therefore we calculated the success rate as $R_{d} / R$ where $R_{d}$ is the drop in recall after the attack, and $R$ is the recall of the base model on the unperturbed data. Table~\ref{Performance} listed the maximum performance for each classifier as found by the grid-search. FIGA achieves a 100\% success rate against logistic regression and the two neural networks with a low epsilon ($<1.0$). FIGA can obtain a $>96\%$ success rate against all the tree models. However, it requires the maximum epsilon. 

\subsection{Success Rate Curves} \label{succ_curves}
While a grid search will yield the argmax for success rate, it does not take into account the \emph{cost} of the perturbation. Intuitively an attack would like to find the ``sweet spot" of the size of perturbation vs. the success rate. In order to find the trade-offs and examine each hyper-parameters effect on the FIGA attack, we plot curves where the attack's success rate is presented on the y-axis, and $n$ and $\epsilon$ are plotted on the x-axis. The success rate plotted is the \emph{maximum} success rate (arg-max corresponding hyper-parameters) found for the selected parameter on the x-axis. In order to examine the impact of $f_{i}$, we plotted bar charts that display the maximum success rate associated with each $f_{i}$.

\begin{figure}
    \centering
    \includegraphics[width=.45\textwidth]{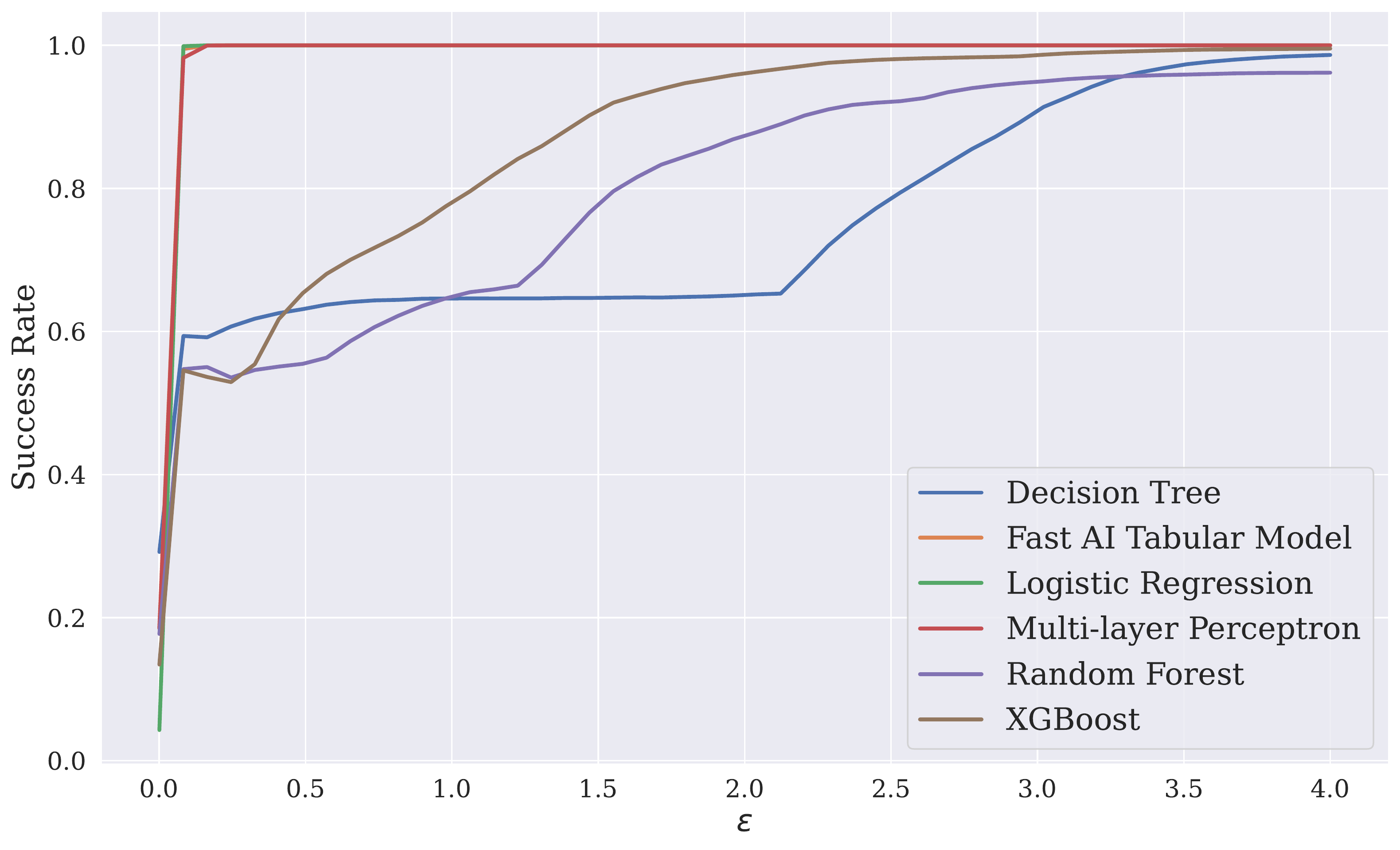}
    \caption{We plot the maximum success rate as we vary $\epsilon$. A clear trend is that the tree based models are more robust to FIGA, required much larger perturbation amounts to achieve higher success rates.}
    \label{fig:FIGA-epsilon}
\end{figure}

Figure~\ref{fig:FIGA-epsilon} demonstrates the maximum success rate as $\epsilon$ varies on the primary dataset. The plot provides empirical evidence that $\epsilon$ positively correlates with the success rate. There is a clear difference between the performance of the tree-based models and the others. The non-tree models, Fast AI tabular model, logistic regression, and MLP are all easily evaded with a low epsilon amount. Among the tree models, the simple decision tree is the most robust to the attack, only reaching the 90\% success rate threshold with $\epsilon >3.0$.

\begin{figure}
\centering
\includegraphics[width=.45\textwidth]{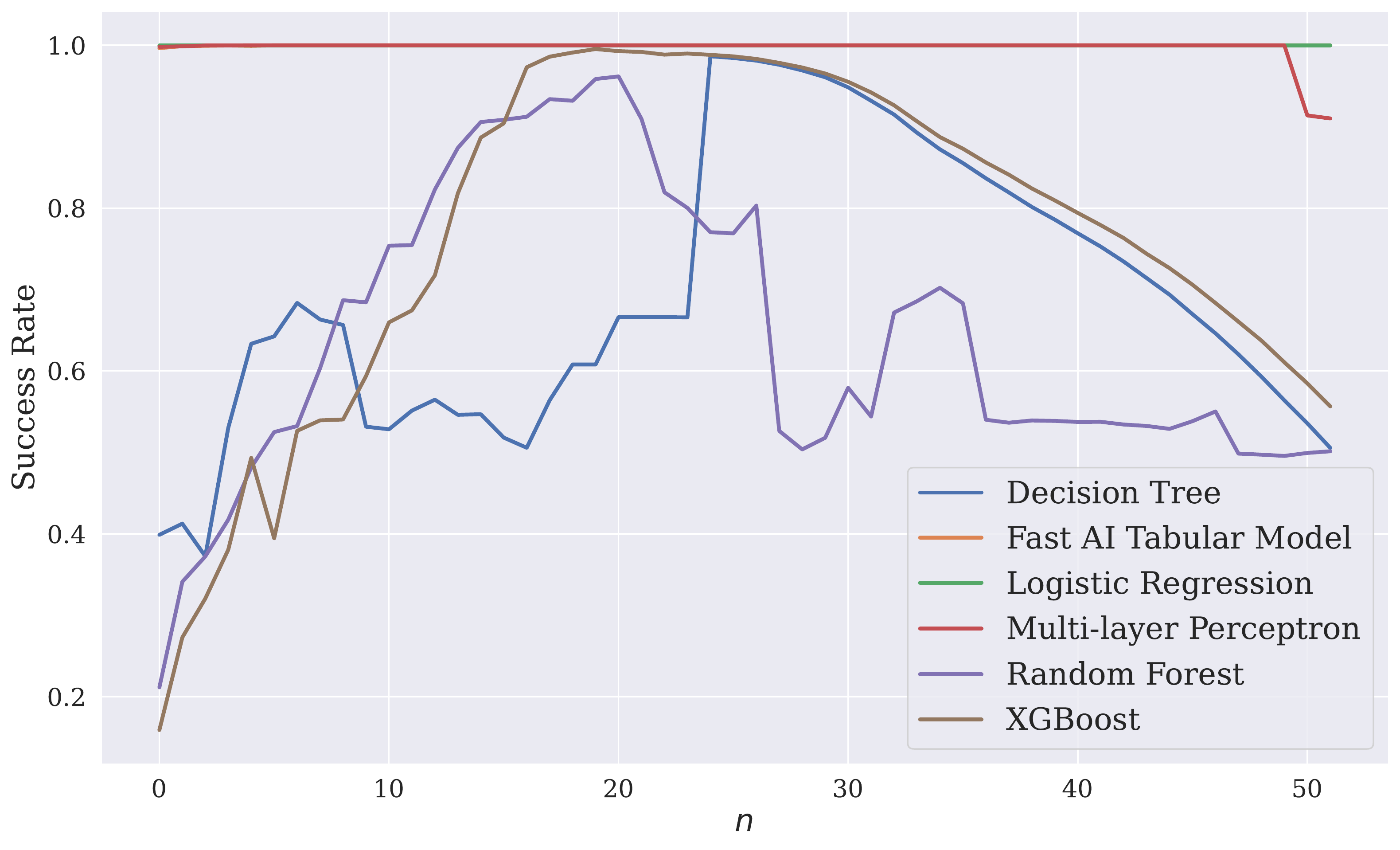}
\caption{We plot the maximum success rate as we vary $n$ on the primary dataset. It is interesting to note that for all values of $n$, we can obtain a 100\% success for the non-tree models. Among the tree-based models, there appear to be individual tipping points of $n$ where the performance peaks. For all three tree models, the optimum $n$ is somewhere in the range of 15-30.}
\label{fig:FIGA-n}
\end{figure}

Figure~\ref{fig:FIGA-n} shows the success rate curve plotted for $n$ ranging from 1 to 52. Again we see a similar separation of the tree vs. non-tree models. The non-tree models can be evaded with $n=1$. The tree models demonstrate some variety - it is clear that there is no $n$, which is universal for all models. For example, when $n$ moves from 22-26, the performance of the attack against decision trees improves, while the performance against random forest decreases. The blue line (decision tree) seems almost to be an inverse of the purple line (random forests). 

\begin{figure}
    \centering
    \includegraphics[width=.45\textwidth]{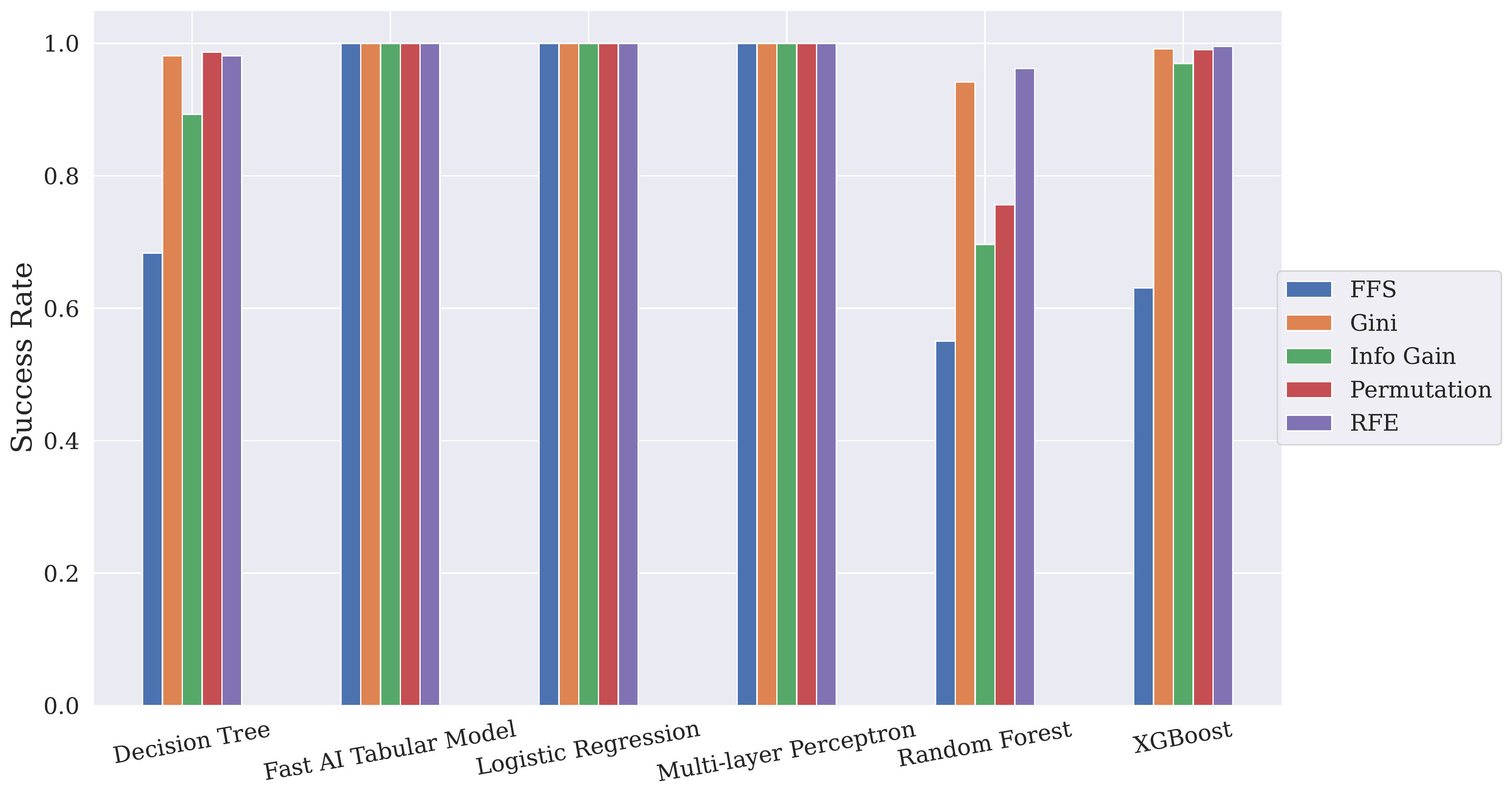}
    \caption{We plot the maximum success rate for each $f_{i}$ algorithm on the primary dataset. Generally, we observe that for the non-tree-based models, all $f_{i}$ choices are equally performant. Both RFE and Gini appear to give the best results within the tree-based models. In general, FFS does not perform well..}
    \label{fig:FIGA-f_i}
\end{figure}

\begin{table*}[tbh!]
\centering
\caption{FIGA performance on the Alternate Datasets in Feature Space}
\begin{tabular}{lccccclc}
\textbf{Dataset}                   & \multicolumn{5}{c}{\textbf{Mamun}}                                                                          & \multicolumn{1}{l}{}      \\ \hline
\multicolumn{1}{c}{Trained Models}  & \multicolumn{2}{c}{Recall Score}  & Success Rate\%         & \emph{n}     & \multicolumn{1}{c}{$\epsilon$} & $f_{i}$                        \\ \hline
                                                       & Baseline & \multicolumn{1}{c}{Attack} &                      & &                      &                           \\ \hline
Fast AI Tabular Model & 97.8 & 0.0 & 100.0 & 60.00 & 2.531 & \emph{Info Gain} \\
Multi-layer Perceptron & 98.3 & 0.0 & 100.0 & 57.00 & 2.776 & \emph{Info Gain} \\
Logistic Regression & 95.1 & 0.0 & 100.0 & 33.00 & 2.449 & \emph{Info Gain} \\
Decision Tree & 97.1 & 1.4 & 98.5 & 32.00 & 4.000 & \emph{Gini} \\
Random Forest & 98.1 & 6.1 & 93.8 & 68.00 & 2.123 & \emph{Permutation} \\
XGBoost & 98.6 & 0.0 & 100.0 & 55.00 & 2.857 & \emph{Info Gain}\\
                                   & \multicolumn{1}{l}{} &          &                            & \multicolumn{1}{l}{} &                       & \multicolumn{1}{l}{}      \\

\textbf{Dataset}                   & \multicolumn{1}{l}{} & \multicolumn{3}{c}{\textbf{Kang}}                            &                       & \multicolumn{1}{l}{}      \\ \hline
\multicolumn{1}{c}{Trained Models}  & \multicolumn{2}{c}{Recall Score}  & Success Rate\%         & \emph{n}                    & \multicolumn{1}{c}{$\epsilon$} & $f_{i}$                        \\ \hline
                                                        & Baseline & \multicolumn{1}{c}{Attack} &                       & &                       &                           \\ \hline
Fast AI Tabular Model & 98.4 & 0.0 & 100.0 & 10.00 & 0.654 & \emph{Info Gain}\\
Multi-layer Perceptron & 97.2 & 0.0 & 100.0 & 38.00 & 2.123 & \emph{FFS}\\
Logistic Regression & 94.3 & 0.0 & 100.0 & 8.00 & 0.491 & \emph{Info Gain}\\
Decision Tree & 96.5 & 0.0 & 100.0 & 13.00 & 1.633 & \emph{Info Gain}\\
Random Forest & 98.0 & 0.0 & 100.0 & 21.00 & 0.899 & \emph{Info Gain}\\
XGBoost & 98.1 & 0.0 & 100.0 & 13.00 & 0.491 & \emph{Info Gain}\\
                                   & \multicolumn{1}{l}{} &          &                            & \multicolumn{1}{l}{} &                       & \multicolumn{1}{l}{}      \\

\textbf{Dataset}                   & \multicolumn{1}{l}{} & \multicolumn{3}{c}{\textbf{Lee}}                             &                       & \multicolumn{1}{l}{}      \\ \hline
\multicolumn{1}{c}{Trained Models} & \multicolumn{2}{c}{Recall Score}   & Success Rate\%            & \emph{n}                    & \multicolumn{1}{c}{$\epsilon$} & $f_{i}$                        \\ \hline
                                   &                       Baseline & \multicolumn{1}{c}{Attack} &                      &                       &       &                    \\ \hline
Fast AI Tabular Model & 86.6 & 0.0 & 100.0 & 8.00 & 0.164 & \emph{Info Gain}\\
Multi-layer Perceptron & 91.2 & 0.0 & 100.0 & 4.00 & 0.164 & \emph{Info Gain}\\
Logistic Regression & 84.9 & 0.0 & 100.0 & 1.00 & 0.083 & \emph{Info Gain}\\
Decision Tree & 93.0 & 1.0 & 98.9 & 26.00 & 1.225 & \emph{Permutation}\\
Random Forest & 94.3 & 0.0 & 100.0 & 32.00 & 3.674 & \emph{Gini}\\
XGBoost & 94.4 & 0.1 & 99.9 & 16.00 & 2.939 & \emph{Permutation}\\
                                   & \multicolumn{1}{l}{} &          &                            &                      &                       & \multicolumn{1}{l}{}      \\
\textbf{Dataset}                   & \multicolumn{1}{l}{} & \multicolumn{3}{c}{\textbf{Adult}}                           &                       & \multicolumn{1}{l}{}      \\ \hline
\multicolumn{1}{c}{Trained Models}  & \multicolumn{2}{c}{Recall Score}  & Success Rate\%         & \emph{n}                    & \multicolumn{1}{c}{$\epsilon$} & $f_{i}$                        \\ \hline
                                   &                       Baseline & \multicolumn{1}{c}{Attack} &                      &                      &  &                           \\ \hline
Fast AI Tabular Model & 90.3 & 0.0 & 100.0 & 9.00 & 0.980 & \emph{Info Gain}\\
Multi-layer Perceptron & 87.7 & 0.0 & 100.0 & 8.00 & 1.062 & \emph{Info Gain}\\
Logistic Regression & 84.1 & 0.0 & 100.0 & 9.00 & 1.144 & \emph{Info Gain}\\
Decision Tree & 87.8 & 0.0 & 100.0 & 10.00 & 1.144 & \emph{Info Gain}\\
Random Forest & 87.9 & 0.0 & 100.0 & 10.00 & 0.736 & \emph{Info Gain}\\
XGBoost & 88.8 & 0.0 & 100.0 & 13.00 & 1.062 & \emph{Info Gain}\\  \bottomrule
\end{tabular}

\label{tab:other_data}
\end{table*}

Figure~\ref{fig:FIGA-f_i}  shows the impact of the feature importance algorithm $f_{i}$ on the FIGA attack. Again we see a clear distinction between the tree and non-tree models, where for the highly successful attacks, it appears that all $f_{i}$ algorithms are equal. For the more robust tree models, RFE and Gini appear more performant.

In general, the findings of the grid search indicate that of the three parameters, $e$ is the only parameter correlated with the success rate. $n$ needs to be tuned per model, as there is often a tipping point where increasing the number of features to perturb yields negative results. Finally, $f_{i}$ is less significant of a factor, with either Gini or RFE appearing to be usable across all models.

\subsection{Performance on the alternate datasets}
\begin{figure*}[!ht]
\begin{multicols}{4}
    \begin{subfigure}{\linewidth}
    \includegraphics[width=\linewidth]{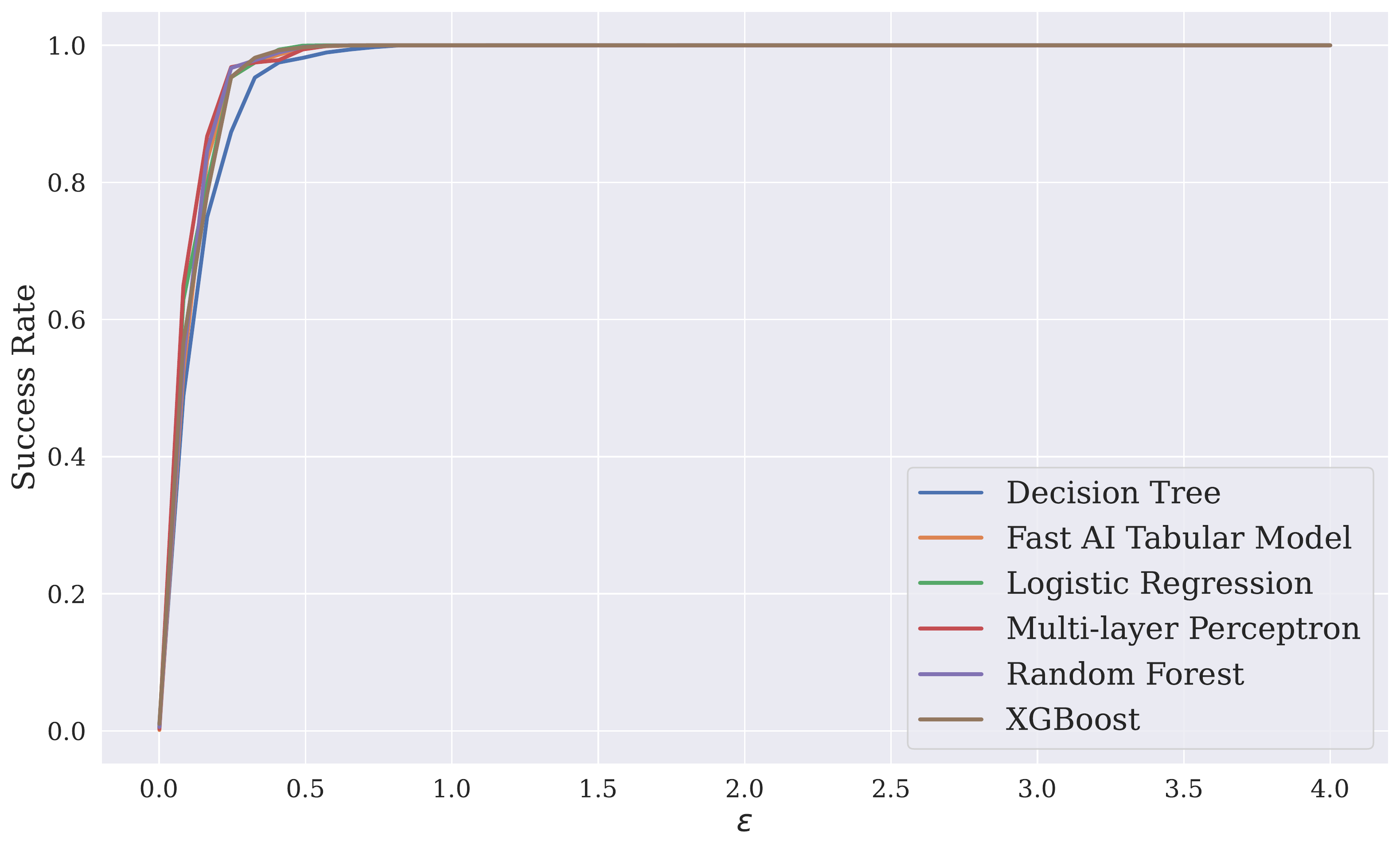}
    \caption{Adult}
    \label{fig:Adult-epsilon}
    \end{subfigure}
    
    \begin{subfigure}{\linewidth}
    \includegraphics[width=\linewidth]{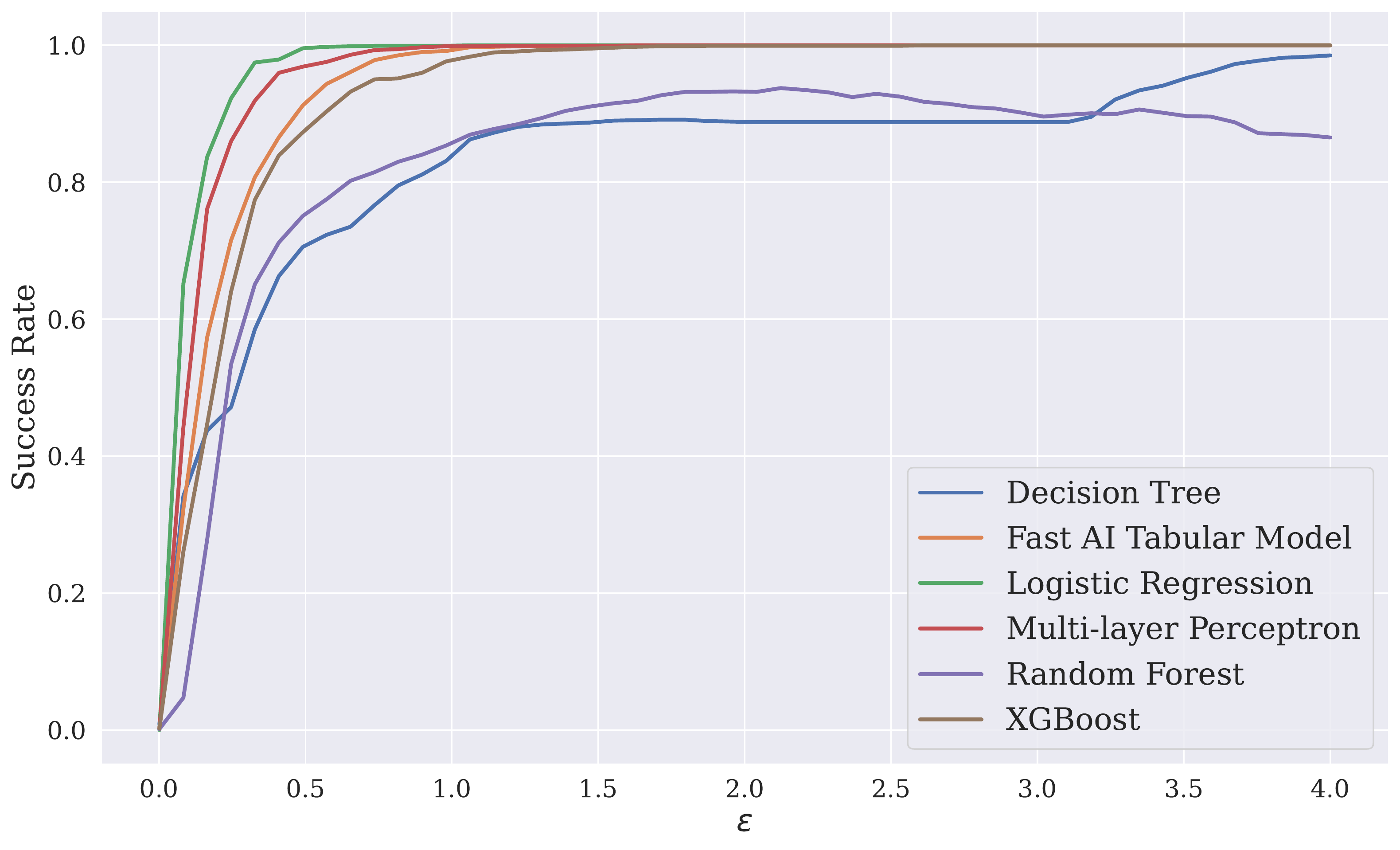}
    \caption{Mamun}
    \label{fig:Mamun-epsilon}
    \end{subfigure}

    \begin{subfigure}{\linewidth}
    \includegraphics[width=\linewidth]{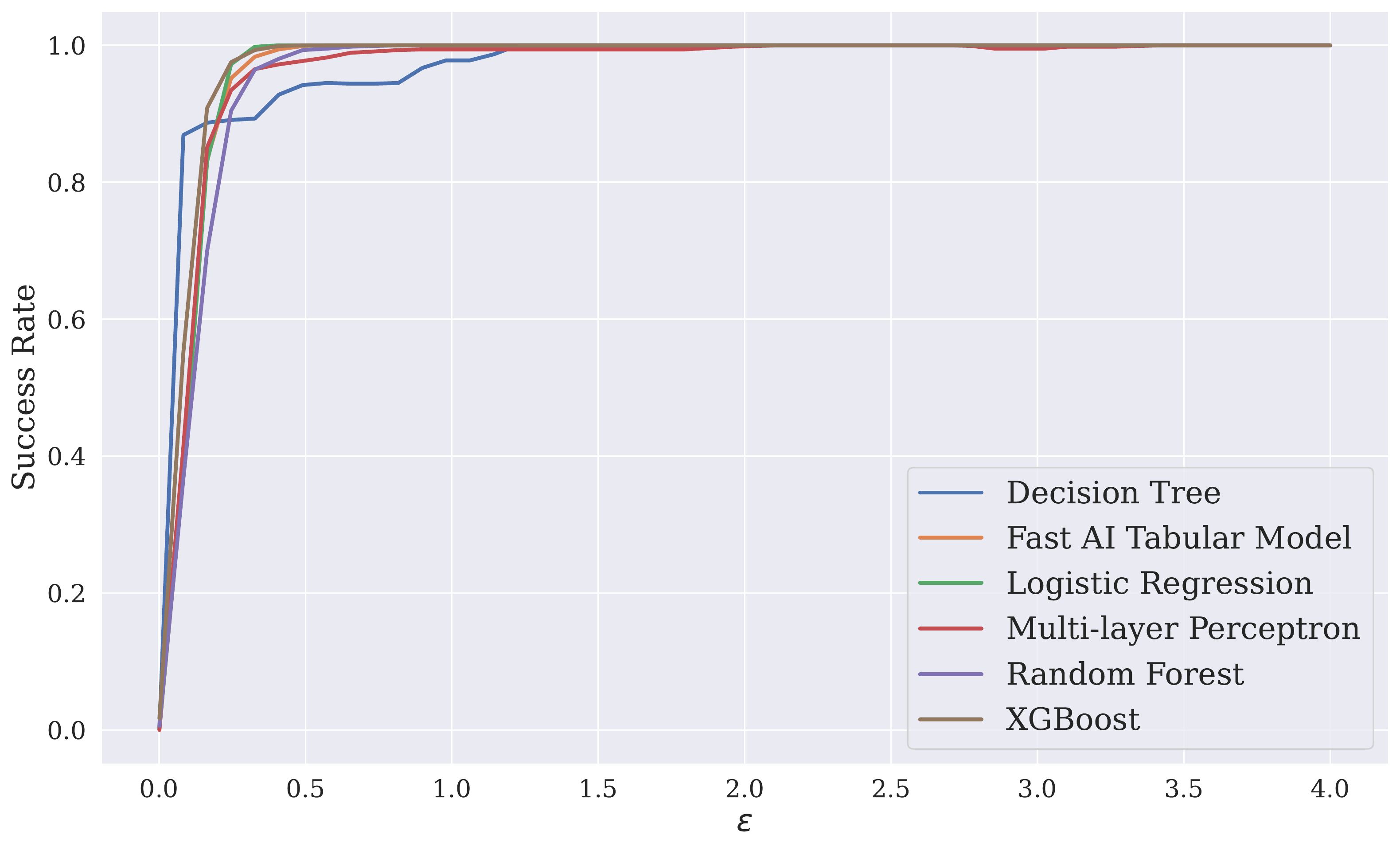}
    \caption{Kang} 
    \label{fig:Kang-epsilon}
    \end{subfigure}
    
    \begin{subfigure}{\linewidth}
    \includegraphics[width=\linewidth]{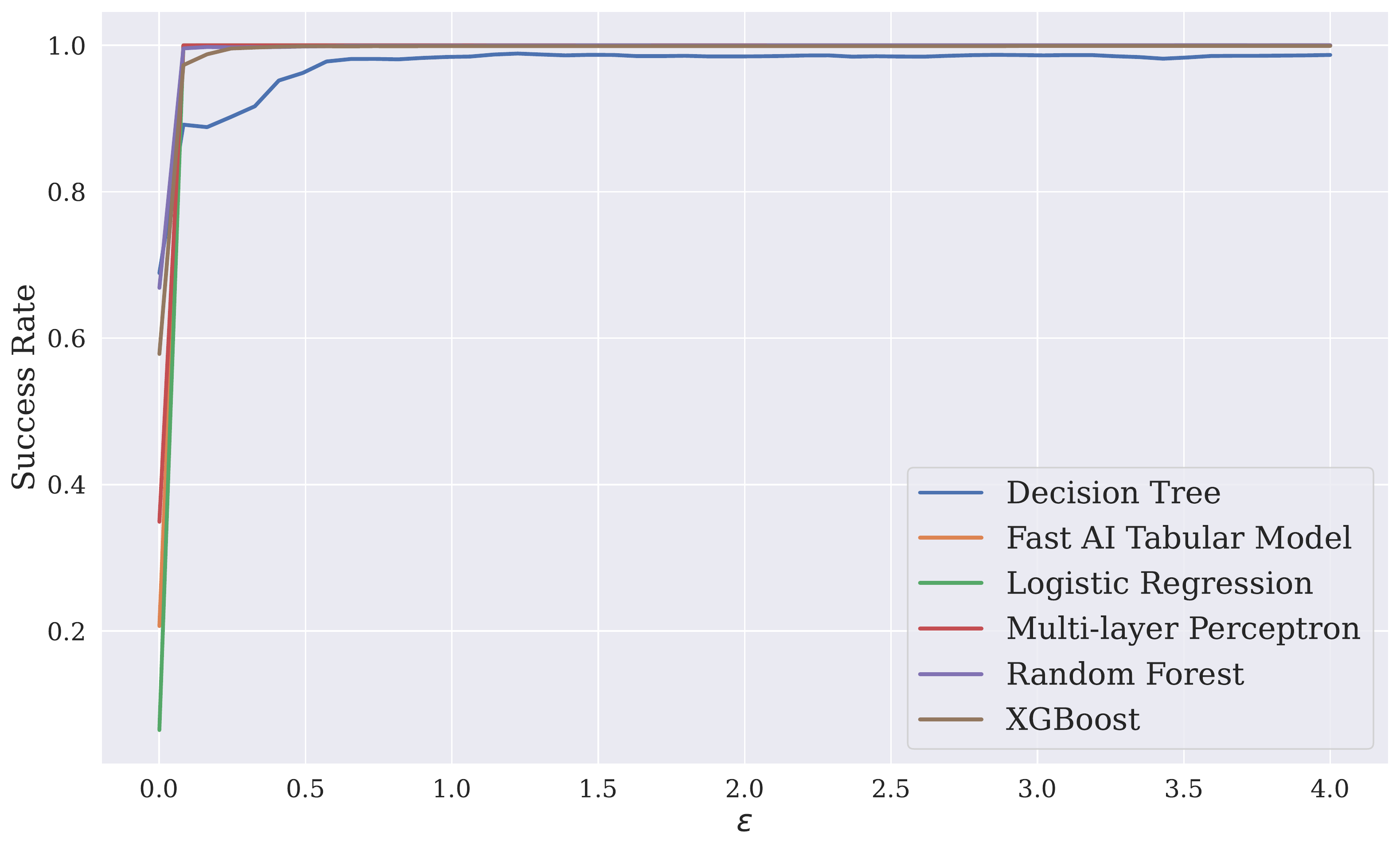}
    \caption{Lee}
    \label{fig:LEE-epsilon}
    \end{subfigure}
\end{multicols}
\caption{The maximum success rate as $\epsilon$ increases for the alternative datasets}
\label{fig: alternate data eps curves}
\end{figure*}

\begin{figure*}[h!]
\begin{multicols}{4}
    \begin{subfigure}{\linewidth}
        \includegraphics[width=\linewidth]{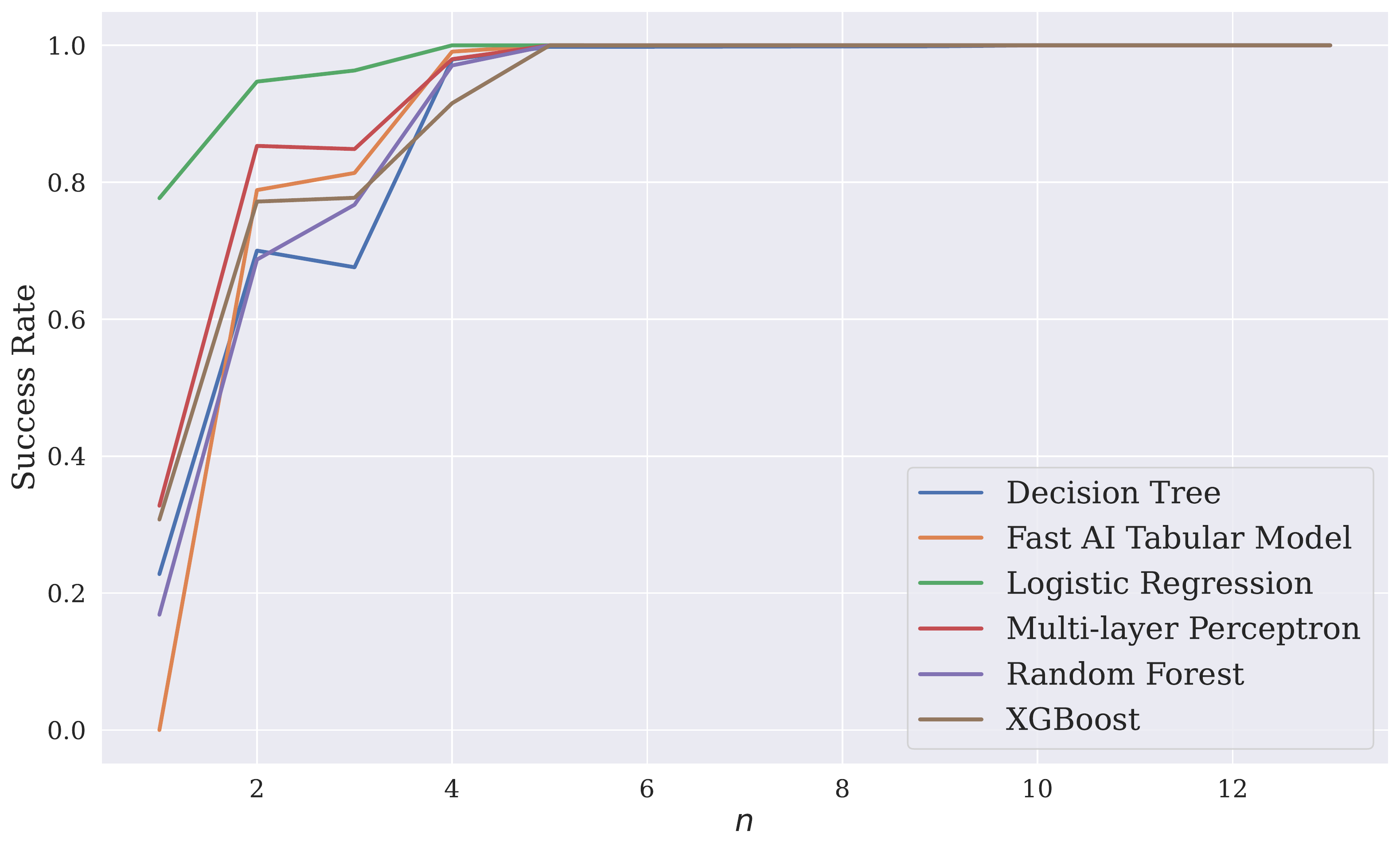}
        \caption{Adult}
        \label{fig:Adult-n_feat}
    \end{subfigure}

    \begin{subfigure}{\linewidth}
        \includegraphics[width=\linewidth]{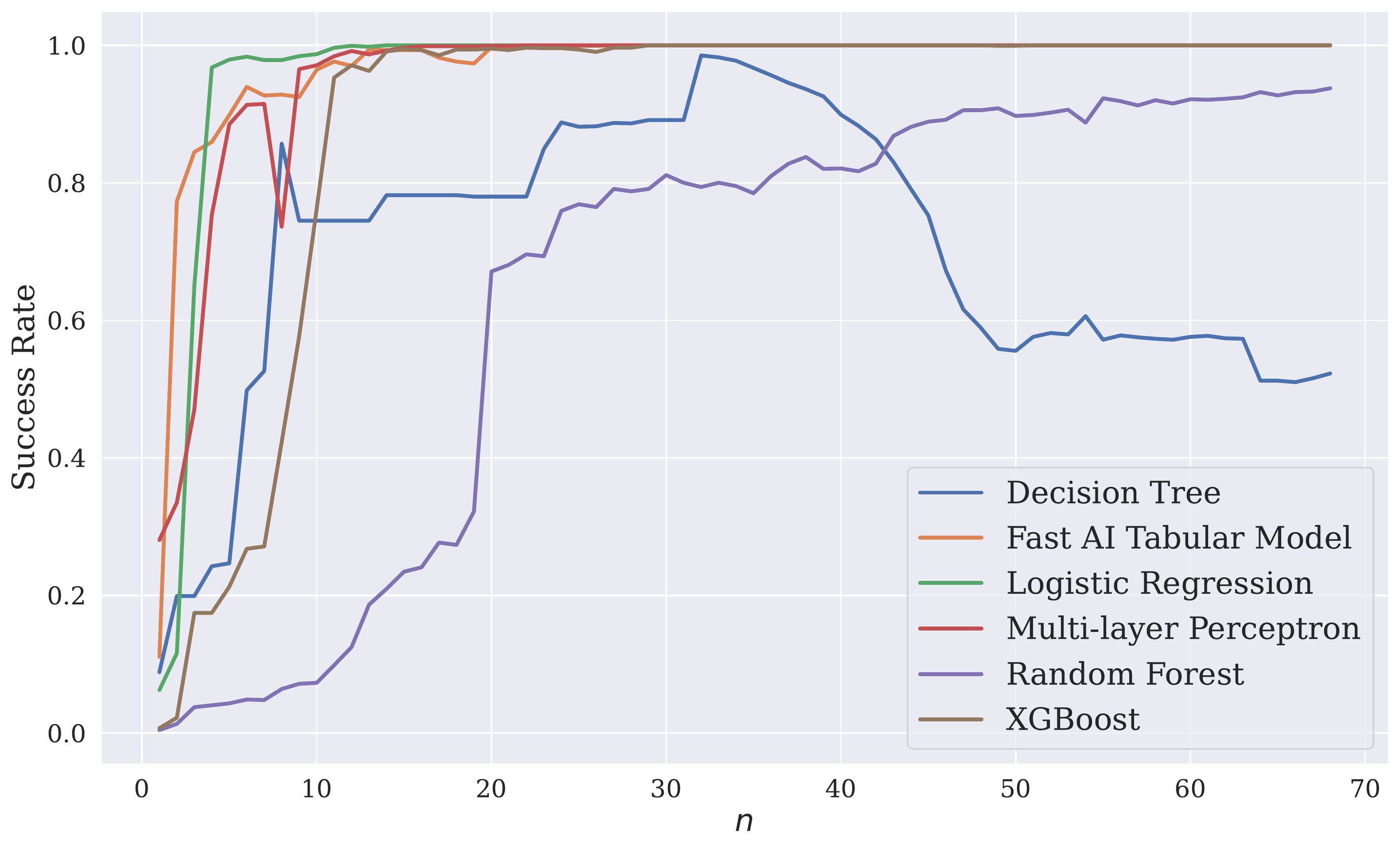}
        \caption{Mamun}
        \label{fig:mamun-n_feat}
    \end{subfigure}

    \begin{subfigure}{\linewidth}
        
        \includegraphics[width=\linewidth]{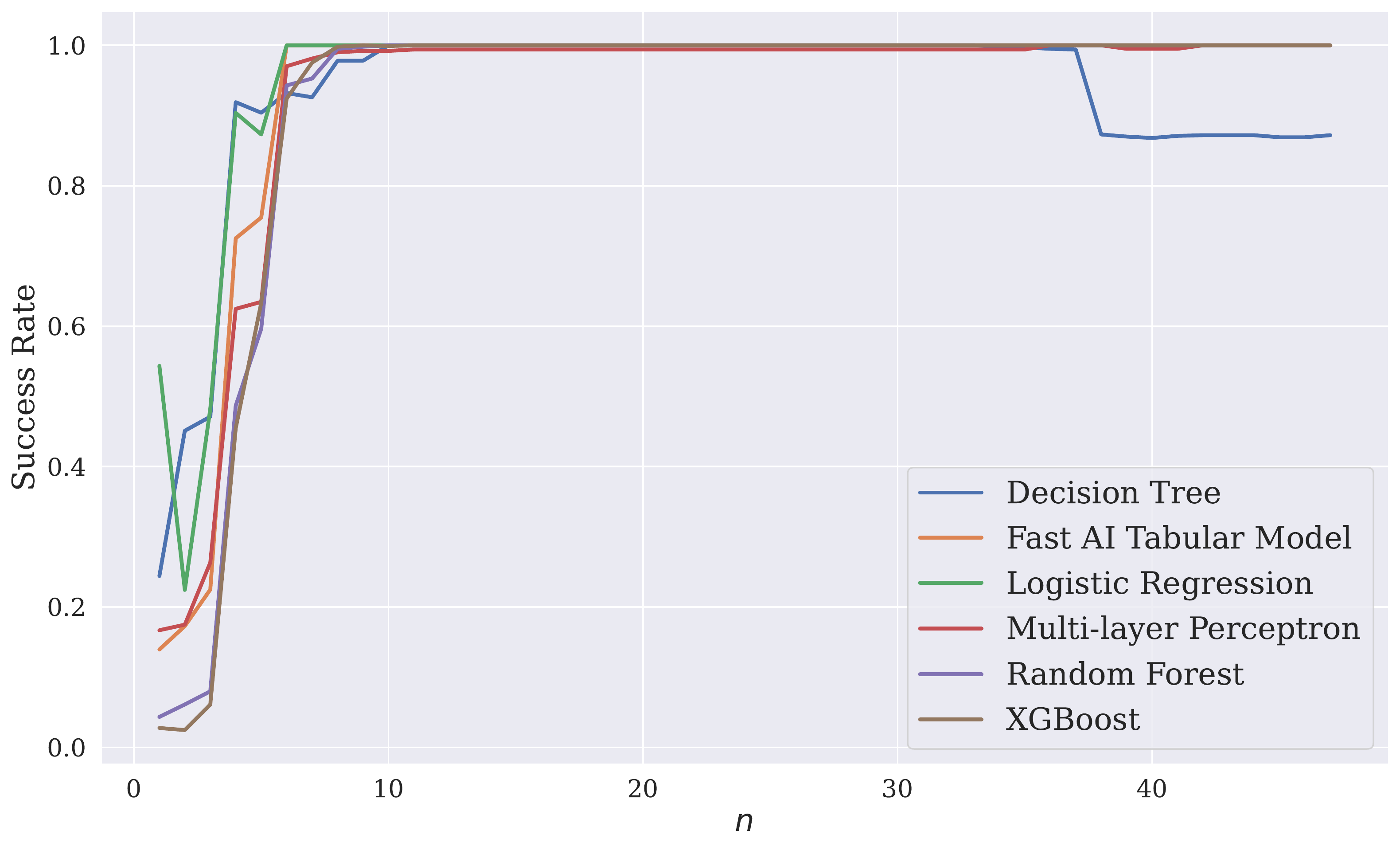}
        \caption{Kang}
        \label{fig:Kang-n_feat}
    \end{subfigure}

    \begin{subfigure}{\linewidth}
        \includegraphics[width=\linewidth]{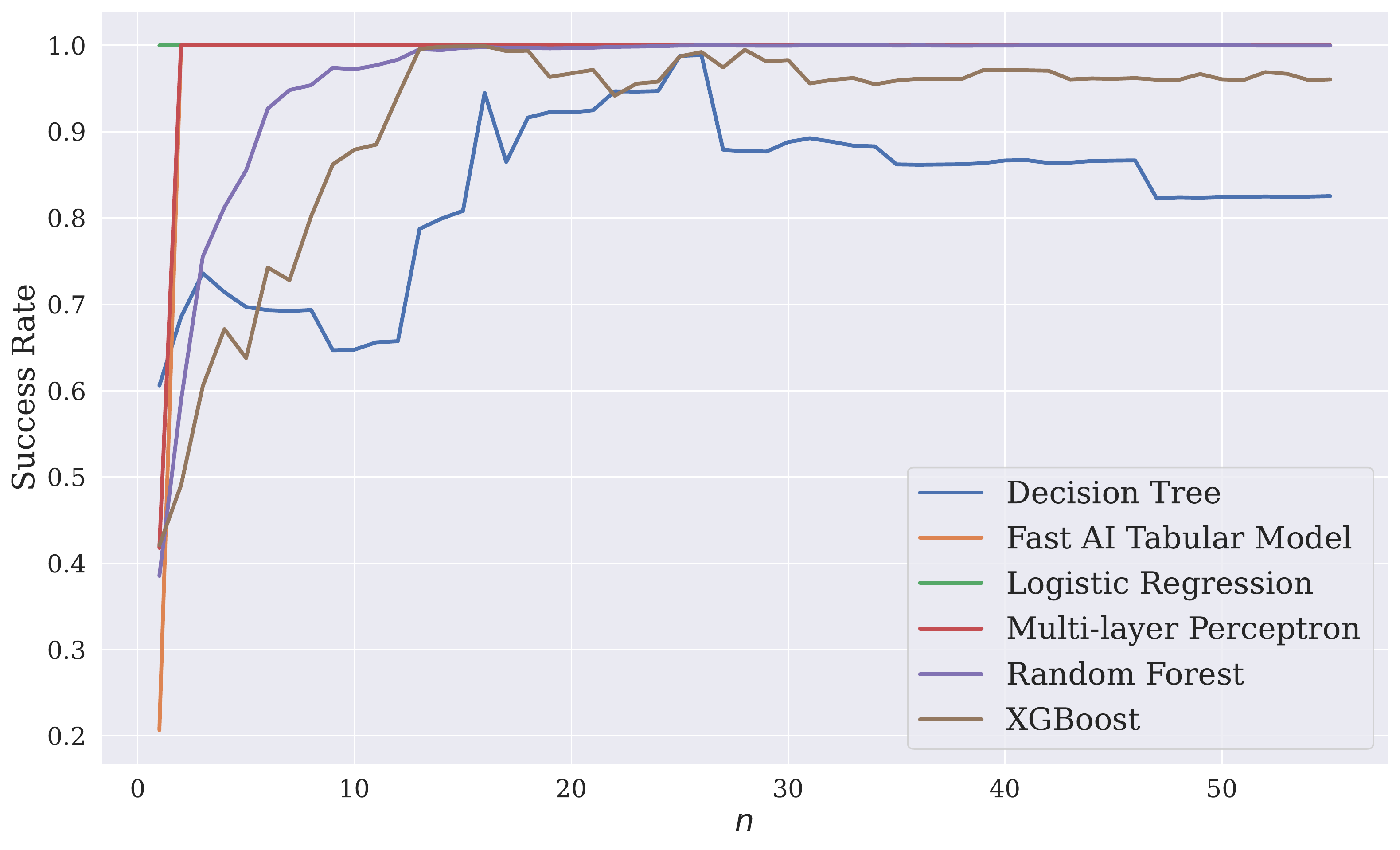}
        \caption{Lee}
        \label{fig:LEE-n_feat}
    \end{subfigure}
\end{multicols}

\caption{The maximum success rate as $n$ increases for all datasets}
\label{fig:alternate_N_range}
\end{figure*}

\begin{figure*}[!ht]
\begin{multicols}{4}
    \begin{subfigure}{\linewidth}
        \includegraphics[width=\linewidth]{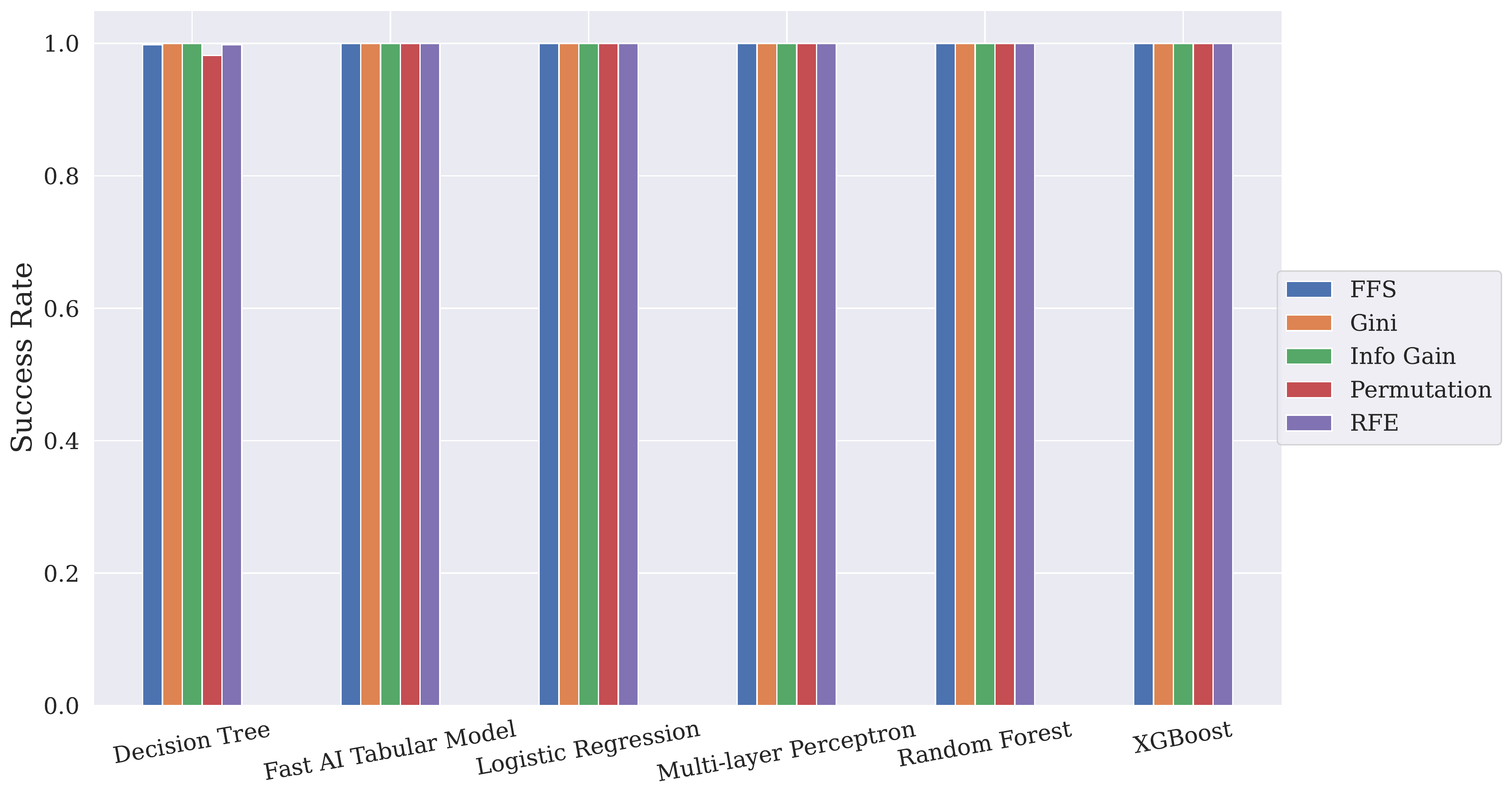}
        \caption{Adult}
        \label{fig:Adult-imp}
    \end{subfigure}

    \begin{subfigure}{\linewidth}
        \includegraphics[width=\linewidth]{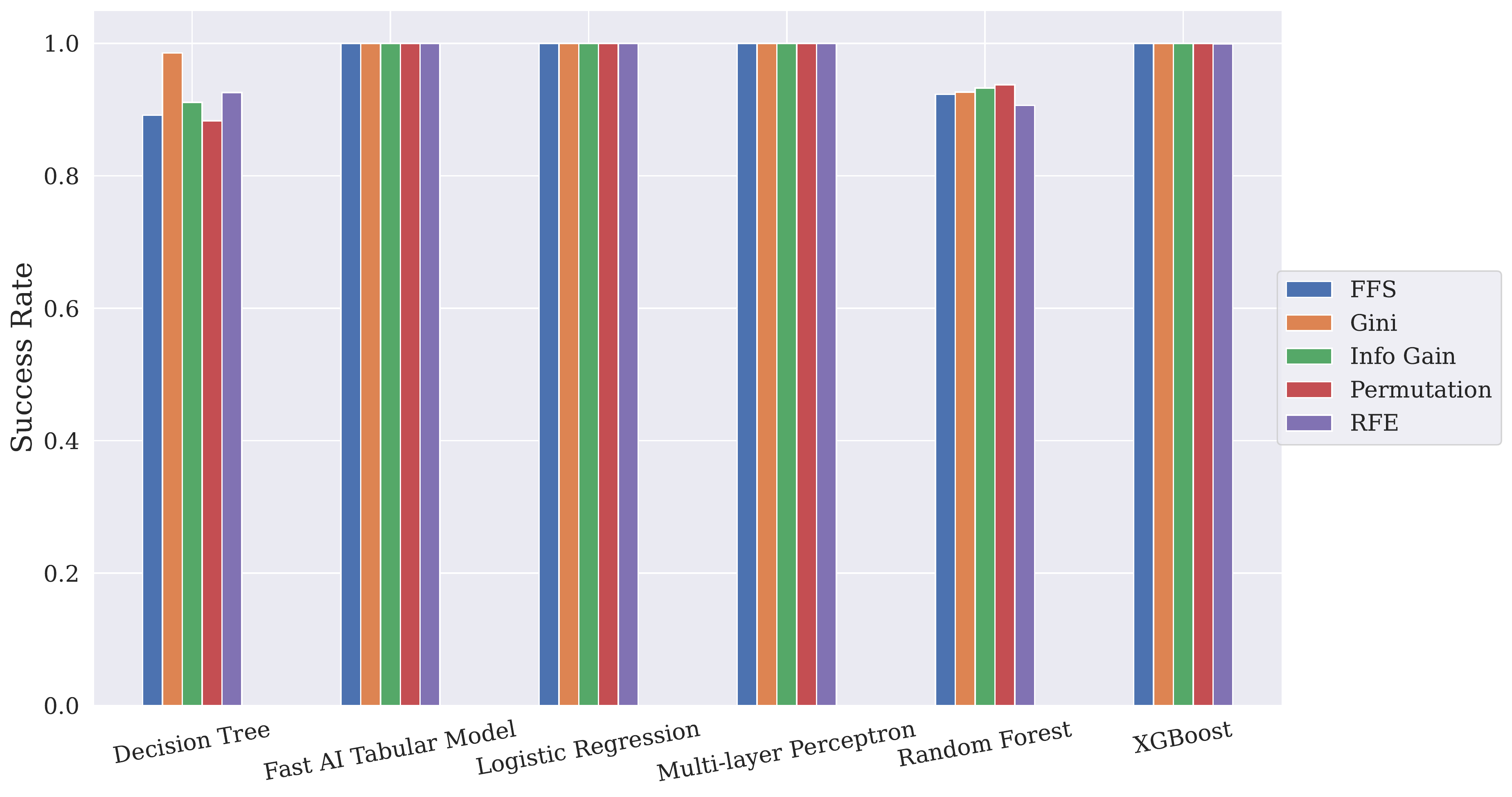}
        \caption{Mamun}
        \label{fig:mamun-imp}
    \end{subfigure}

    \begin{subfigure}{\linewidth}
        
        \includegraphics[width=\linewidth]{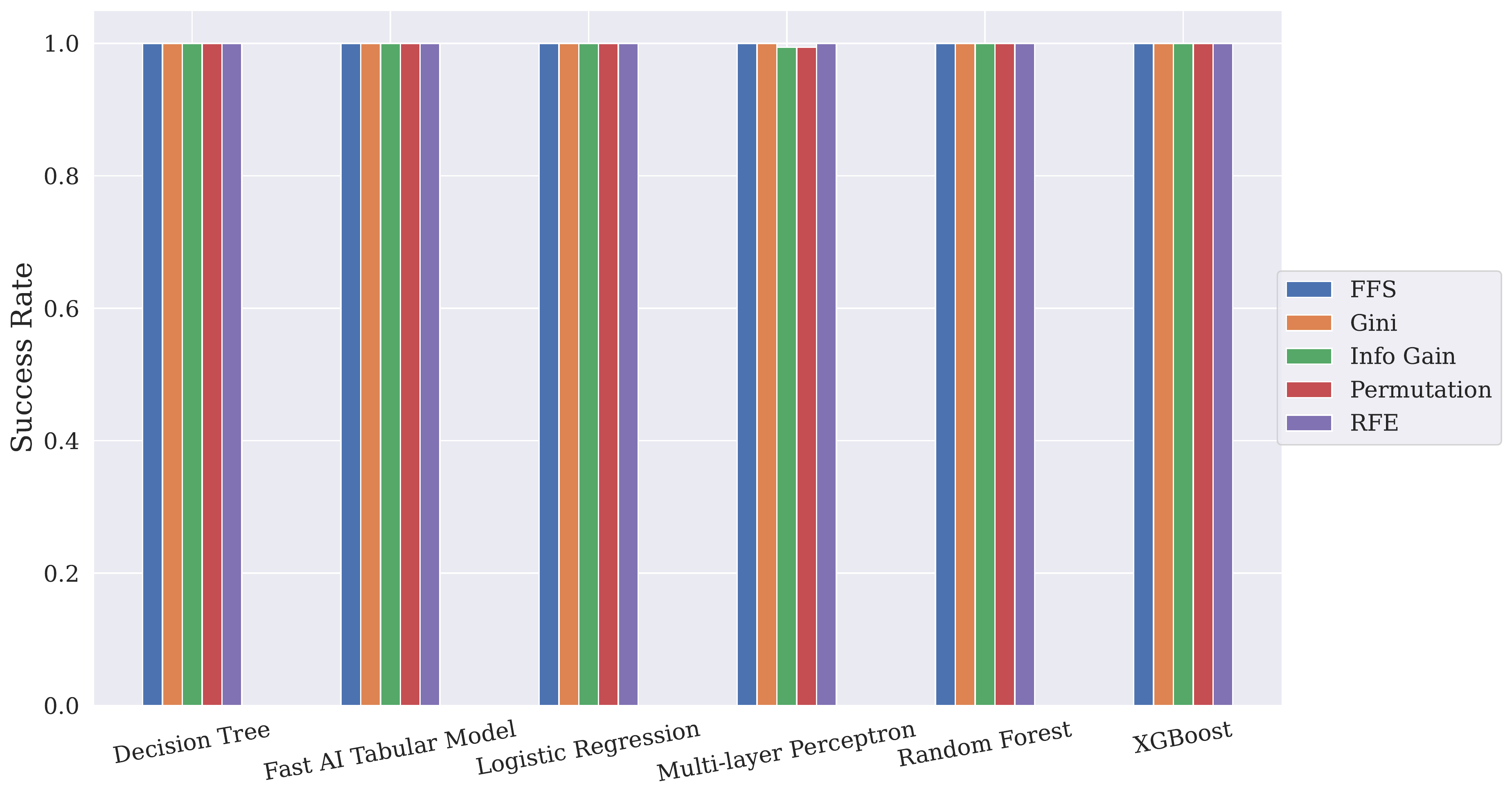}
        \caption{Kang}
        \label{fig:Kang-imp}
    \end{subfigure}

    \begin{subfigure}{\linewidth}
        \includegraphics[width=\linewidth]{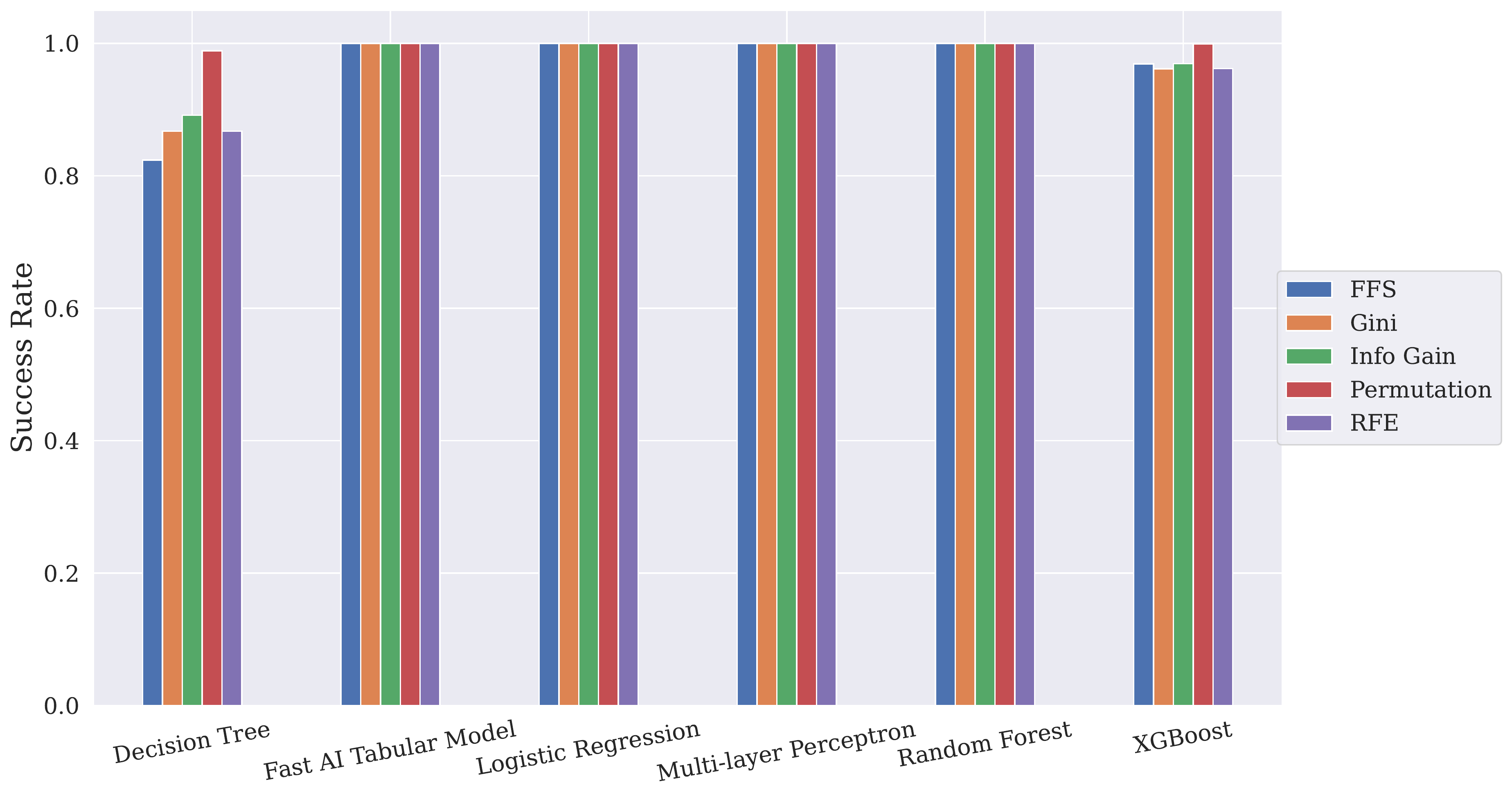}
        \caption{Lee}
        \label{fig:LEE-imp}
    \end{subfigure}
\end{multicols}

\caption{Maximum Success Rates for all $f_{i}$ algorithms.}
\label{fig:alternate_imps}
\end{figure*}

In Table~\ref{tab:other_data}, we display the success rate of FIGA on the alternate datasets. In general, FIGA performs more efficiently against the additional datasets than when compared to the primary dataset. The required $\epsilon$ is lower, and the number of features needed is also less. Figure~\ref{fig: alternate data eps curves} shows a success curve as $\epsilon$ varies; we notice similar patterns with the Mamun dataset, where decision trees and random forests are more robust against the FIGA attack. Figure~\ref{fig:alternate_N_range} demonstrates the effect of $n$ on the success rate. The Adult and Kang datasets can be evaded with very low $n$ while Mamun and Lee require higher $n$ for the tree algorithms, similar to the primary dataset results. Figure ~\ref{fig:alternate_imps} shows bar charts for all $f_{i}$ options, as with the primary dataset, a maximum attack is possible with all $f_{i}$ algorithms. Overall we observe that FIGA performs consistently across all datasets.


\section{Problem Space Attack - Primary Dataset}

\begin{table}[!ht]
\centering
\caption{Features used to generate adversarial samples in problem space}
\begin{tabular}{@{}lc@{}}
\toprule
\textbf{Feature}  & \textbf{Sign}   \\ 
\midrule
\# href           & +               \\
\# javascript     & +               \\
\# text in body   & +               \\
\# images         & +               \\
\# meta tag       & +               \\
\# forms          & +                \\
\# iframes         & +                \\
\# hidden text    & +  \\
\# redirects     & +  \\
\# submit to mail & + \\
\bottomrule
\end{tabular}
\label{table:prob_space_feat}
\end{table}

\begin{table*}
\centering
\begin{tabular}{lccccclc}
\toprule
Trained Models      & \multicolumn{2}{c}{Recall}  & Success Rate\% & $n$  & $\epsilon$     & $f_{i}$ \\ \toprule
                       & Baseline      & Attack          &           &         &           &                           \\ \hline
FastAI Tabular Model    & 94.5         & 74.6.    & 21.09    & 5 & 4.0 & \emph{gini}         \\ 
Multi Layer Perceptron  & 97.7         & 85.3     & 12.69    & 5  & 4.0 & \emph{gini}                      \\
Logistic Regression     & 87..4         & 86.2     & 1.36     & 8  & 6.0 & \emph{gini}                      \\
Decision Tree           & 94.7         & 72.0     & 24.0    & 8 & 6.0 & \emph{gini}                      \\
XGBoost                 & 98.6         &  90.2     & 8.49    & 8  & 6.0 & \emph{gini}               \\
Random Forest           & 97.7         & 87.2     & 10.72     & 8 & 6.0 & \emph{gini}        \\ \bottomrule  

\end{tabular}
\caption{Performance in Problem Space}
\label{problem_space_performance}
\end{table*}

In order to create a problem space attack, we must convert the feature space perturbation found by FIGA into actual HTML / URL source code. We can only perform the problem space attack on the primary dataset as this is the only phishing dataset that contains source code (the rest have only shared the extracted features). In order to simplify the process and guarantee functional websites, we chose to only \emph{add} features in the HTML code. However, we are confident that the URL could also be modified; we leave that for future work. We do not subtract any features from the HTML, which could break the existing website's functionality. Of the 52 available features, only 22 are HTML features, and 21 have a positive sign. However, not all of these features can be repeatedly perturbed since they are often binary categorical. For example, protocol (HTTPS vs. HTTP) and title (does the website have a title) are two binary features that cannot be repeated. Due to memory constraints we had to remove text\_in\_body, the perturbations on that feature where creating enormous (gb) websites, which were both not practical or realistic. After selecting only positive features that can be repeatedly perturbed, we arrived at a list of 8 features listed in Table~\ref{table:prob_space_feat}.  

\begin{figure*}
\begin{multicols}{2}
    \begin{subfigure}{\linewidth}
        \includegraphics[width=\linewidth]{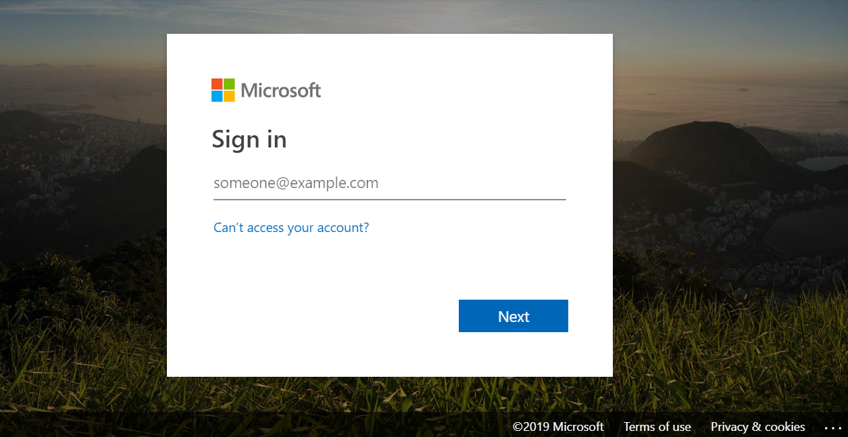}
        \caption{Before}
    \end{subfigure}

    \begin{subfigure}{\linewidth}
        \includegraphics[width=\linewidth]{images/Picture1.png}
        \caption{After}
    \end{subfigure}

\end{multicols}
\caption{A phishing webpage before and after problem space attack. We manually checked 100 pages to ensure that all pages loaded correctly and were visually identical to their unperturbed counterparts.}
\label{visuall_simiar}
\end{figure*}

We performed a grid-search with FIGA varying $\epsilon$ from 0.001 to 8.0 and $n$ from 1-8, allowing FIGA only to search the eight features we can easily inverse map back into problem space. We limited the feature importance ranking algorithm to Gini impurity since earlier results indicated that the importance ranking does not make a significant difference. From the feature space search, we select the best $\epsilon$ and $n$ and then inverse the attack into problem space.
We use the BeautifulSoup~\cite{richardson2007beautiful} library to automatically generate all the HTML elements required by the FIGA attack. In Figure~\ref{visuall_simiar}, we present a website before and after perturbation. We manually opened 100 websites in a browser to ensure that the problem space version both loaded correctly and was visually identical to the unperturbed version.

Once the attack is in problem space, we re-run the original feature extraction. This is required because of the side-effect features generated unintentionally. For example, when we add images, we will naturally increase the body's text. After re-running the feature extraction, we pass the adversarial samples to our trained models and observe the success rate. The feature text\_in\_body would increase naturally as a side effect of adding other features. However, side effects can also reduce the mimicry of perturbation. 

The problem space results are presented in Table~\ref{problem_space_performance}. The problem space attack is \emph{not} as effective as the feature space attack. However, we would argue that the problem space attack could be vastly improved with more focused and clever engineering. Certainly, modifying the URL is plausible, and so is removing elements from websites. Still, the average success rate of 13.05\% would be attractive to potential attackers who build phishing websites in mass.

\section{Conclusion and Future Work}
This work presents the novel `Feature Importance Guided Attack (FIGA)' effective against tabular datasets. FIGA is a simple mimicry attack that does not do any complex optimizations. However, it is highly effective on tabular datasets in feature space and can be used as a search strategy to find perturbations to apply in problem space.

We thoroughly studied the three parameters involved in the FIGA attack and determined that $n$ and $\epsilon$ are far more critical than $f_{i}$ the ranking algorithm. We extended FIGA to the problem space, demonstrating the first (to our knowledge) problem space adversarial phishing websites.

In future work, we would like to examine the minimum amount of knowledge FIGA needs to perform successful attacks. Currently, we make use of the entire training data to learn the mean statistics in order to perform the mimicry perturbations. However, we believe that this is not required. A minimum sampling could likely find the direction of the attack quite quickly. Finally, FIGA is currently a "1-shot" attack that does not perform iterative steps, and we would also consider modifying this to ensure a higher success rate, especially in the problem space.

\bibliographystyle{IEEEtran}
\bibliography{IEEEabrv,bib}

\appendix
\section{Appendix}
\label{appendix}
\setcounter{table}{0}
\begin{table*}
\centering
\caption{Feature Rank list with their sign,type and description}
\begin{tabularx}{\linewidth}{|c|c|c|c|X|} 
\toprule
Rank & Feature                  & Sign & Type & Description                                                                                                                   \\ 
\hline
1    & href                     & +    & HTML & Number of href tags                                                                                                           \\ 
\hline
2    & javascript               & +    & HTML & Number of Javascript                                                                                                          \\ 
\hline
3    & text\_in\_body           & +    & HTML & Number of words                                                                                                               \\ 
\hline
4    & no\_www                 & +    & URL  & Number of `www'~                                                                                                              \\ 
\hline
5    & images                   & +    & HTML & Number of Images~                                                                                                             \\ 
\hline
6    & meta                     & +    & HTML & Number of meta tags                                                                                                           \\ 
\hline
7    & no\_digits               & -    & URL  & Number of digits~                                                                                                             \\ 
\hline
8    & subdomain\_len           & -    & URL  & Length of the subdomain                                                                                                       \\ 
\hline
9    & alph\_digit\_ratio       & +    & URL  & Ratio of alphabets and digits~                                                                                                \\ 
\hline
10   & url\_len                 & -    & URL  & Length of the URL                                                                                                             \\ 
\hline
11   & len\_freeurl             & -    & URL  & Length of the URL after removing the protocol part                                                                            \\ 
\hline
12   & no\_dir                  & -    & URL  & Number of paths~                                                                                                              \\ 
\hline
13   & no\_alphanumeric         & -    & URL  & Number of alphanumeric characters                                                                                             \\ 
\hline
14   & hyphens\_in\_path           & -    & URL  & Number of ``-" characters~                                                                                      \\ 
\hline
15   & longest\_token           & -    & URL  & Length of longest word~                                                                                                       \\ 
\hline
16   & suspicious\_words        & +    & HTML & Count of words such as `cardnumber',`cvv', `email', `submit', `prepaid', `bitcoin', `log in', `sign up', `logon', `register'  \\ 
\hline
17   & len\_fqdn                & -    & URL  & Length of the URL except protocol and ``/" characters.                                                          \\ 
\hline
18   & protocol                 & +    & URL  & Binary feature encoding HTTP as 0 and HTTPS as 1                                                                              \\ 
\hline
19   & passwdfield              & -    & HTML & Number of password field in website                                                                                           \\ 
\hline
20   & no\_vowels               & -    & URL  & Number of vowels~                                                                                                             \\ 
\hline
21   & no\_alpha                & -    & URL  & Number of alphabets~                                                                                                          \\ 
\hline
22   & no\_constants            & -    & URL  & Number of consonants~                                                                                                         \\ 
\hline
23   & no\_dots                 & -    & URL  & Number of dots~                                                                                                               \\ 
\hline
24   & host\_dig\_let\_ratio    & +    & URL  & Ratio of alphabets and digits in URL excluding protocol and paths                                                             \\ 
\hline
25   & iframes                  & +    & HTML & Number of iframes tag                                                                                                         \\ 
\hline
26   & forms                    & +    & HTML & Number of forms tag                                                                                                           \\ 
\hline
27   & length\_of\_domains      & -    & URL  & Length of domain~                                                                                                             \\ 
\hline
28   & dots\_freeurl            & -    & URL  & Number of dots in only domain                                                                                                 \\ 
\hline
29   & relativeforms            & +    & HTML & Number of form action tag                                                                                                     \\ 
\hline
30   & vowel\_constant\_ratio   & +    & URL  & Ratio of vowel and consonant~                                                                                                 \\ 
\hline
31   & hidden\_text             & +    & HTML & Number of hidden text tag                                                                                                     \\ 
\hline
32   & longest\_token\_hostname & -    & URL  & Length of longest word in hostname                                                                                            \\ 
\hline
33   & dig\_in\_hostname        & -    & URL  & Length of hostname~                                                                                                           \\ 
\hline
34   & no\_dash                 & -    & URL  & Number of ``-" character~                                                                                                      \\ 
\hline
35   & redirects                & +    & HTML & Number of redirects using javascript                                                                                          \\ 
\hline
36   & url\_of\_anchor          & +    & HTML & Number of anchor tags                                                                                                         \\ 
\hline
37   & submit\_to\_mail         & +    & HTML & Number of `href=mailto' tag                                                                                                   \\ 
\hline
38   & rightclick\_disabled     & +    & HTML & Number of rightclick disabled tags                                                                                            \\ 
\hline
39   & no\_special\_sym         & -    & URL  & Number of special symbols such as @,\#~                                                                                       \\ 
\hline
40   & title                    & +    & HTML & If title tag present, then 1. Else 0                                                                                          \\ 
\hline
41   & no\_percent              & -    & URL  & Number of percentage sign characters~                                                                                         \\ 
\hline
42   & no\_eq                   & -    & URL  & Number of equal sign characters~                                                                                              \\ 
\hline
43   & no\_ques                 & -    & URL  & Number of question mark character~                                                                                            \\ 
\hline
44   & popup                    & +    & HTML & Number of popup in website                                                                                                    \\ 
\hline
45   & insecureforms            & +    & HTML & Number of forms tag using http protocol                                                                                       \\ 
\hline
46   & no\_http                 & -    & URL  & Number of ``http"~                                                                                                             \\ 
\hline
47   & abnormalforms            & +    & HTML & Number of abnormal forms~                                                                                                     \\ 
\hline
48   & onmouseover              & +    & HTML & Presence of onmouseover tag~                                                                                                  \\ 
\hline
49   & no\_at                   & -    & URL  & Number of @ characters~                                                                                                       \\ 
\hline
50   & userprompt               & +    & HTML & Number of prompt tag~                                                                                                         \\ 
\hline
51   & no\_dollar               & -    & URL  & Number of \$ character~                                                                                                       \\ 
\hline
52   & SFH                      & +    & HTML & Number of safe forms~                                                                                                         \\

\hline

\multicolumn{5}{|r|}{} \\ \hline
\end{tabularx}
\end{table*}

\end{document}